\documentclass[conference]{IEEEtran}
\IEEEoverridecommandlockouts
\usepackage{hyperref}
\usepackage{graphicx}
\usepackage{tikz}
\usetikzlibrary{positioning, arrows.meta, fit, backgrounds}
\usepackage{multirow}
\usepackage{array}
\usepackage{enumitem}
\usepackage{url}
\usepackage{threeparttable}
\usepackage{amsmath,amssymb,amsbsy,xspace}

\makeatletter
\DeclareRobustCommand\onedot{\futurelet\@let@token\@onedot}
\def\@onedot{\ifx\@let@token.\else.\null\fi\xspace}

\def\figmulti{0.8}
\DeclareRobustCommand*{\IEEEauthorrefmark}[1]{\raisebox{0pt}[0pt][0pt]{\textsuperscript{\footnotesize #1}}}
\makeatother

\begin{document}

\title{Representation Recycling for Streaming Video Analysis\\
\thanks{\textsuperscript{*}Equal contribution for senior authorship.}
}

\author{%
\IEEEauthorblockN{Can Ufuk Ertenli\IEEEauthorrefmark{1},
Ramazan Gokberk Cinbis\textsuperscript{1,2*},
Emre Akbas\textsuperscript{1,2*}}
\IEEEauthorblockA{\IEEEauthorrefmark{1}Department of Computer Engineering, Middle East Technical University (METU), Ankara, Turkey}
\IEEEauthorblockA{\IEEEauthorrefmark{2}Center for Robotics and Artificial Intelligence (ROMER), Middle East Technical University (METU), Ankara, Turkey}
}

\maketitle

\begin{abstract}
We present StreamDEQ, a method that aims to infer frame-wise representations on videos with minimal per-frame computation. Conventional deep networks perform feature extraction from scratch at each frame in the absence of ad-hoc solutions. We instead aim to build streaming recognition models that can natively exploit temporal smoothness between consecutive video frames. We observe that the recently emerging implicit layer models provide a convenient foundation to construct such models, as they define representations as the fixed points of shallow networks, which need to be estimated using iterative methods. Our main insight is to distribute the inference iterations over the temporal axis by using the most recent representation as a starting point at each frame. This scheme effectively recycles the recent inference computations and greatly reduces the required processing time. Through extensive experimental analysis, we show that StreamDEQ is able to recover near-optimal representations within a few frames and maintain an up-to-date representation throughout the video duration. Our experiments on video semantic segmentation, video object detection, and human pose estimation in videos show that StreamDEQ achieves on-par accuracy with the baseline while providing 2-4x higher throughput.
\end{abstract}

\begin{IEEEkeywords}
Implicit layer models, streaming video understanding, object detection, semantic segmentation, human pose estimation, deep equilibrium models.
\end{IEEEkeywords}

\section{Introduction}
\label{sec:intro}

Modern convolutional deep networks excel at numerous recognition tasks. It is commonly observed that deeper models tend to outperform their shallower counterparts~\cite{he2016deep,huang2017densely}. Due to the sequential dependencies across the layers, however, increasing the network depth results in longer processing times. While the increase in inference duration can be acceptable for various offline recognition problems, it is typically of concern for many streaming video analysis tasks. For example, in perception modules of autonomous systems, it is not only necessary to keep up with the frame rate but also desirable to minimize the computational burden of each recognition component to reduce the hardware requirements~\cite{yang2022real}. Similarly, in large-scale video analysis tasks, small changes in per-frame computations can lead to significant cumulative cost differences.

\subsection{Limitations of Existing Video Acceleration Techniques}

\begin{figure}
    \centering
    \includegraphics[width=0.49\textwidth]{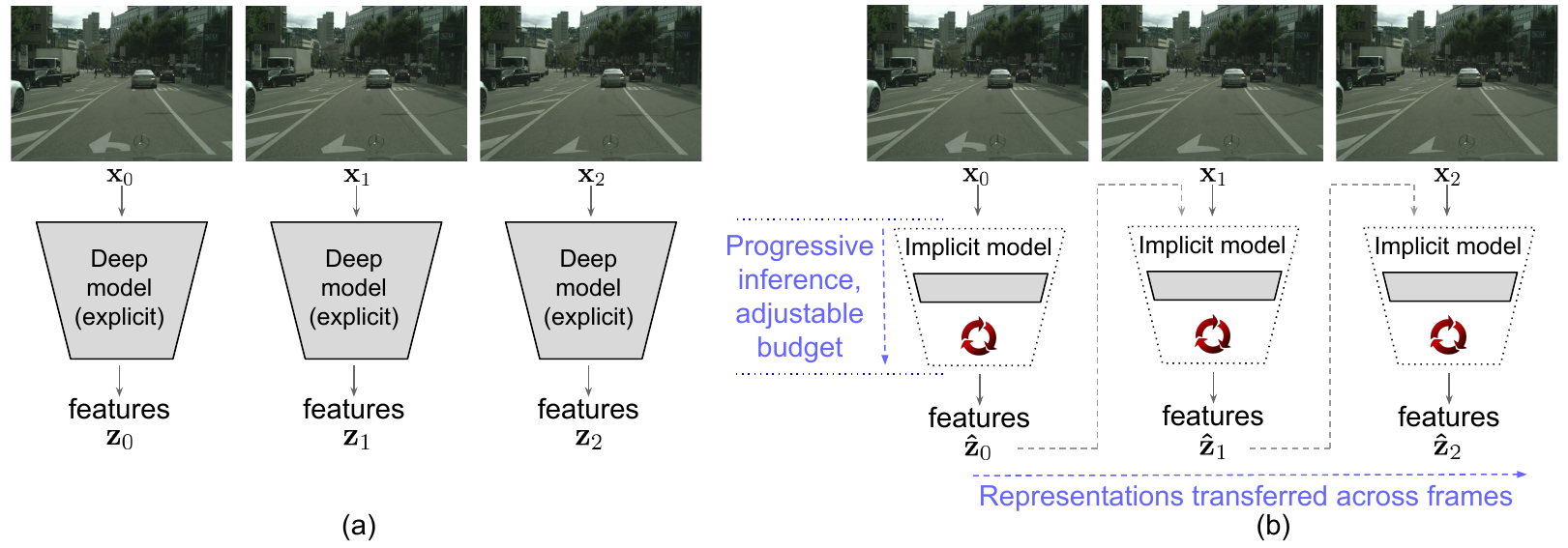}  
    \caption{(a) Naive per-frame processing. (b) The proposed scheme allows reusing features continuously during the stream and dynamically adjusting the number of inference iterations, where at each time step the current frame's representation progressively improves over the iterations.}
    \label{fig:teaser}  
\end{figure}

The naive approach to video analysis applies deep networks independently to each frame (Fig.~\ref{fig:teaser}a). While easy to implement, this is highly inefficient because it ignores the inherent temporal smoothness of videos, yet are ubiquitous in industrial applications and many academic works as well~\cite{zhou2022transvod,wang2022ptseformer,duan2022revisiting,ponimatkin2023simple}. To speed up inference, alternative techniques often apply a large model only to selected {\em key frames}. They then interpolate features to intermediate frames~\cite{zhu2017dff,zhu2018towards} or apply smaller secondary models~\cite{xu2018dynamic,liu2019looking}. However, such approaches introduce several complications. These include the computational overhead of motion estimators~\cite{zhu2017dff,zhu2018towards}, the challenge of maintaining feature compatibility across frames, and a strong dependency on annotated video training data. It is also noteworthy that several models, e.g.~\cite{kang2017tubelets,wang2018manet}, rely on forward \textbf{and} backward flow estimates, making them less suitable for streaming recognition problems due to non-causal processing.

\subsection{Proposed Approach: StreamDEQ}

Instead of relying on explicit optical flow or skipping frames, we envision a model that natively exploits the temporal smoothness\footnote{We do not imply a mathematical definition of smoothness and only emphasize that the differences between neighboring frames are small.} of videos. The ideal system should recycle features from previous frames to avoid from-scratch computation, while progressively improving representation quality as the compute budget allows (Fig.~\ref{fig:teaser}b). These two requirements jointly open doors to important flexibilities for real-time and large-scale systems. For instance, the system may take an action based on early inference results (e.g., emergency braking in an autonomous vehicle), dynamically change compute time based on the instantaneous availability of compute resources, or reduce compute costs on scenes with quick representation convergence.

To achieve this, we leverage the power of implicit layer models, such as the Deep Equilibrium Model (DEQ)~\cite{bai2019deq} and Multiscale DEQ (MDEQ)~\cite{bai2020mdeq}. Unlike conventional deep networks, DEQs define representations as the fixed points of a single shallow network block, estimated via iterative root-finding methods (e.g., Broyden’s method~\cite{broyden1965class}). One can also leverage fixed point iterations by repeatedly applying the same layer to its output to again find the fixed point, which would be analogous to the rolled-out version of DEQ. Because each iteration effectively acts as an increment in network depth, DEQs offer a flexible computational budget. However, standard DEQs initialize these iterations from scratch, leading to high run-time costs that are unsuitable for streaming video.

Our main insight is that the inference process can be drastically accelerated by exploiting temporal smoothness across neighboring frames. Because consecutive frames are visually similar, their fixed points are near each other in the feature space. Therefore, we propose StreamDEQ: instead of fully estimating the representation at each time step from scratch, we use the most recent frame's representation as a warm-start initialization. By running only a few inference iterations per frame, StreamDEQ continuously updates and accumulates the representation across time.

\begin{figure*}
\centering
\includegraphics[width=\linewidth]{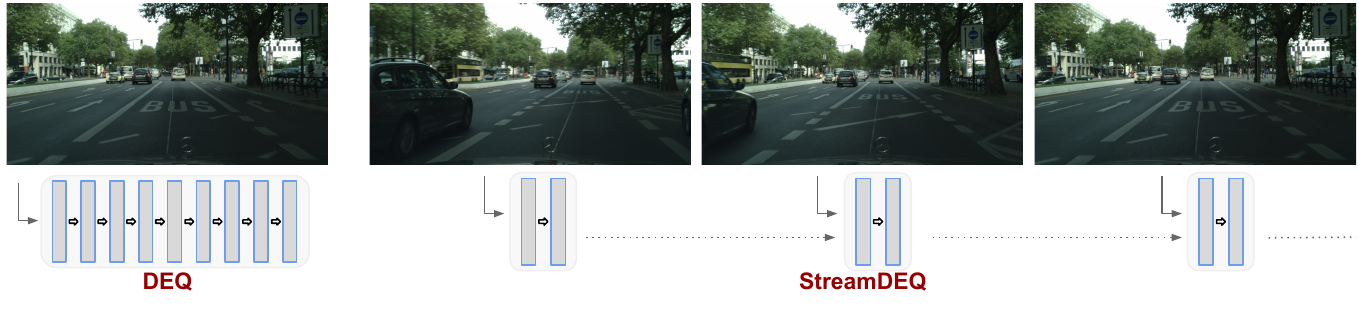}
\caption{Our method, StreamDEQ, exploits the temporal smoothness between successive frames and extracts features via a small number of solver iterations, starting from the previous frame's representation as initial solution. StreamDEQ accumulates and transfers the extracted information continuously over successive frames; effectively sharing computations across video frames in a causal manner.}
\label{fig:intro_figure}
\end{figure*}

The main difference between standard DEQ and our StreamDEQ is illustrated in Fig.~\ref{fig:intro_figure}. While DEQ typically requires a large number of inference iterations, StreamDEQ enables inference with only a few iterations per frame by leveraging the relevance of the most recent frame's representation. At the start of a new video stream, or after a major content change (e.g. a shot change), StreamDEQ quickly adapts to the video in a few frames, much like a person adapting her/his focus and attention when watching a new video. In the following frames, it continuously updates the representation to adapt to minor changes (e.g. objects moving, entering, or exiting the scene). We further introduce a stochastic version of StreamDEQ, where the number of iterations per frame is not fixed and determined randomly at each frame. We show that this stochastic version outperforms all other versions of StreamDEQ, i.e. the versions using Broyden's method or unrollings of the model in every downstream task. 

Overall, StreamDEQ provides a simple and {\em lean} solution to streaming recognition both with implicit layer models and the explicit versions of said implicit models, where a single model naturally performs cost-effective recognition without relying on external inputs and heuristics, such as optical flow~\cite{horn1981determining}, post-processing methods (Seq-NMS~\cite{han2016seq} or tubelet re-scoring~\cite{kang2016object,kang2017tubelets}). In contrast to the common requirements of RNN-based video models, e.g.~\cite{lu2017online}, StreamDEQ does not require fixing the number of iterations during training and does not even require (costly-to-annotate) video training data, thanks to the equilibrium-based scheme. Our method also maintains the causality of the system and executes in a continuous manner. We also note that it allows dynamic time budgeting; the duration of the inference process can be tuned on-the-fly by a controller, depending on the instantaneous compute system load. This flexibility is highly desirable in real-world deployments, sharing conceptual similarities with adaptive mechanisms used to maintain and assess quality in modern video streaming infrastructures~\cite{zhou2022brief,amirpour2025vqm4has}.

We verify the effectiveness of the proposed method through extensive experiments on video semantic segmentation, video object detection and human pose estimation in videos. Our experimental results show that StreamDEQ recovers near-optimal representations at much lower inference costs. More specifically, on the ImageNet-VID video object detection task, StreamDEQ converges to the mAP scores of $69.5$ and $70.3$ using only $4$ and $8$ inference iterations per frame, respectively. In comparison, the standard DEQ inference scheme yields only $8.2$ and $32.6$ mAP scores using $4$ and $8$ iterations, respectively. Similarly, on the Cityscapes semantic segmentation task, using StreamDEQ instead of the standard DEQ inference scheme improves the converged streaming mIoU score from $42.3$ to $77.9$ when $4$ inference iterations are used per frame and from $73.2$ to $80.0$ when $8$ iterations are used per frame. Finally, for human pose estimation in videos using the MPII dataset, we achieve $89.4$ and $89.9$ PCKh with StreamDEQ, whereas the standard DEQ inference scheme achieves $42.7$ and $79.5$ PCKh when we perform $4$ and $8$ iterations, respectively. 

\subsection{Contributions and Journal Extensions}

Our contributions are as follows:

\begin{enumerate}[label=(\roman*)]
    \item We introduce a new inference scheme based on DEQ that can be applied to streaming videos to improve processing speeds drastically while maintaining a high level of performance.
    \item To the best of our knowledge, we provide the first attempt at adapting implicit layer models to streaming videos. We achieve this by taking the fixed-point root-finding formulation—traditionally computed from scratch for static images—and extending it across the temporal axis. 
    \item We empirically show that the fixed points for neighboring video frames are near each other; therefore, previous frames' fixed points provide a good starting point for the upcoming frames.
    \item In addition to applying the implicit layers to streaming videos, we also show the effectiveness of our method beyond implicit models in more general terms based on the idea of feature re-use.
    \item We propose two new variations based on DEQ to estimate the fixed point of the current frame using unrollings of the network where the number of unrollings can either be fixed or selected stochastically during training.
\end{enumerate}

This paper is an extension of our previous work~\cite{ertenli2022streaming}. This journal version has been accepted for publication in \emph{Neurocomputing}. Firstly, we extend the inference scheme of StreamDEQ beyond iterative root-finding methods and incorporate fixed point iteration based unrolling scheme. This version of StreamDEQ, which we call Unrolled StreamDEQ or shortly UR-StreamDEQ (Section~\ref{sec:unrolled_deq}), is much faster and more accurate in all our experiments offering speed-ups of up to $4 \times$ while also providing accuracy gains of up to $5 \times$ in some settings. We also propose a stochastic variant of Unrolled StreamDEQ, namely, Stochastically Unrolled StreamDEQ (SUR-StreamDEQ), where the number of iterations is randomly selected on-the-fly during training. We show that this version of StreamDEQ further improves the converged performance in fewer frames. Importantly, we provide insight into why the StreamDEQ scheme works by showing empirical evidence by adding the new Section~\ref{sec:intuition} confirming our intuition. Overall, this extended version presents more than $100$ extra experiments over our previous work including completely new experiments for existing tasks on the unrolled StreamDEQ variants. Furthermore, we added an entirely new section (Section~\ref{sec:pose_est}) which describes our experiments on human pose estimation, as a new downstream task. Additionally, we provide an extended literature review to cover recent developments especially in the field of implicit layer models (Section~\ref{sec:related_implicit}) as well as a new section on human pose estimation in videos (Section~\ref{sec:related_pose_est}). To avoid confusion, from here onward, we refer to the StreamDEQ version using iterative root-finding methods as Implicit StreamDEQ (IL-StreamDEQ).

In the rest of the paper, we first provide an overview of related work in Section~\ref{sec:related_work} and introduce the background in Section~\ref{sec:background}. In Section~\ref{sec:method}, we describe our inference scheme and how we apply StreamDEQ to streaming videos. In Section~\ref{sec:experiments}, we present the details of our experimental setup and demonstrate the results of our extensive experiments to verify the effectiveness of our method on challenging datasets. Finally, we conclude with a summary of our work in Section~\ref{sec:conc}.

\section{Related Work}\label{sec:related_work}

Here, we summarize efficient video processing methods, video object detection, segmentation, and human pose estimation models. Furthermore, we discuss saliency-based techniques for image and video processing. Finally, we overview implicit models and introduce some important application areas.

\subsection{Efficient Video Processing and Inference}

There have been many efforts to improve video processing efficiency to reach real-time processing speeds. Most of these works take a system-oriented approach~\cite{carreira2018massively,narayanan2019pipedream,li2020towards}. For example, Carreira et al.~\cite{carreira2018massively} develop an efficient parallelization scheme over multiple GPUs and process different parts of a model in separate GPUs to improve efficiency while sacrificing accuracy due to frame delays. Narayanan et al.~\cite{narayanan2019pipedream} propose a novel scheduling mechanism that efficiently schedules and divides forward and backward passes over multiple GPUs. In another work, Li et al.~\cite{li2020towards} use a dynamic scheduler that skips frame(s) when the delays build up to the point where it would be impossible to calculate the results of the next frame in the allotted time.

We note that works on low-cost network designs, such as MobileNets~\cite{howard2017mobilenets,sandler2018mobilenetv2} and low-resolution networks~\cite{liu2018mobile,zhao2018deep}, are also relevant. Such efforts are valuable primarily for replacing network components with more compute-friendly counterparts. However, the advantages of such techniques can also be limited due to natural trade-offs between speed and performance~\cite{zhu2020review}, as the lower-cost network components tend to have lower expressive power. Nevertheless, one can easily incorporate low-cost model design principles into DEQ or StreamDEQ models, thanks to the architecture-agnostic definition of implicit layer models. While such efforts may bring reductions in inference wall-clock time, they are outside the scope of our work.

\subsection{Video Semantic Segmentation}

Semantic segmentation is a costly, spatially dense prediction task. Due to this specific nature of the task, its application to videos remains relatively limited. Most works rely on exploiting temporal relations between frames using methods such as feature warping~\cite{gadde2017semantic,xu2018dynamic,huang2018efficient,jain2019accel}, feature propagation~\cite{shelhamer2016clocknet,li2018low,liang2022ant}, feature aggregation~\cite{hu2020temporally}, and knowledge distillation~\cite{liu2020efficient,liu2022efficient,habibian2022delta} to reduce the computational cost. 

Gadde et al.~\cite{gadde2017semantic} propose warping features of the previous frame at different depths based on optical flow. Xu et al.~\cite{xu2018dynamic} evaluate regions of the input frame and decide whether to warp the features with a cheap flow network or use the large segmentation model based on a confidence score. Huang et al.~\cite{huang2018efficient} keep a moving average over time by combining the segmentation maps from the current frame with the warped map from the previous frame. Jain et al.~\cite{jain2019accel} warp high-quality features from the last key frame and fuse them with lower-quality features calculated on the current frame to make predictions.

Shelhamer et al.~\cite{shelhamer2016clocknet} propose an adaptive method that schedules updates to the multi-level feature map so that features of layers with smaller changes are carried forward (without any transformation). Li et al.~\cite{li2018low} introduce an adaptive key frame scheduling method based on the deviation of low-level features compared to the previous key frame. If the deviation is small, the features are propagated with spatially variant convolution. Liang et al.~\cite{liang2022ant} apply the full model on key frames and propose to adaptively choose the depth of the network at each non-key frame.

Hu et al.~\cite{hu2020temporally} use a set of shallow networks, each calculating features of consecutive frames starting from scratch. Then, these features are aggregated at the current frame with an attention-based module. Liu et al.~\cite{liu2020efficient,liu2022efficient} propose to use an expensive network during training and applies knowledge distillation on a student network to cut the test-time compute costs. On the other hand, Habibian et al.~\cite{habibian2022delta} use a teacher network on key frames, and through knowledge distillation, guide the student network to learn feature differences that are combined with the key frame features to simulate the teacher on non-key frames.

In contrast to all these approaches, the proposed StreamDEQ scheme directly leverages the similarities across video frames, without requiring any ad-hoc video handling strategies, as a way to adapt the implicit layer inference mechanism to efficient streaming video analysis.

\subsection{Video Object Detection}

Most modern video object detection methods exploit temporal information to improve the accuracy and efficiency. To this end, optical flow~\cite{zhu2017dff,zhu2017fgfa,kang2017tubelets,wang2018manet}, feature aggregation~\cite{zhu2017fgfa,bertasius2018object,wu2019selsa,chen2020mega} and post-processing techniques~\cite{han2016seq,kang2016object} are prominently used. 

Zhu et al.~\cite{zhu2017dff} introduce Deep Feature Flow (DFF) and use optical flow estimates to warp features on selected key frames to intermediate frames for increased efficiency. Zhu et al.~\cite{zhu2017fgfa} also propose Flow-Guided Feature Aggregation (FGFA) which uses optical flow to warp features of nearby frames to the current frame and aggregates these features adaptively based on feature similarity. Kang et al.~\cite{kang2017tubelets} create links between objects through time (tubelets) from the predictions calculated with optical flow across a video linking objects through time and apply tubelet re-scoring to keep detections of high confidence. Wang et al.~\cite{wang2018manet} add an instance level calibration module to FGFA~\cite{zhu2017fgfa} and combine them to generate better predictions. 

Bertasius et al.~\cite{bertasius2018object} sample features from neighboring support frames via deformable convolution that learns object offsets between frames and aggregates these features over these frames. Wu et al.~\cite{wu2019selsa} focus on linking object proposals in a video according to their semantic similarities. Chen et al.~\cite{chen2020mega} propose a model aggregating local and global information with a long-range memory.

Another common way to improve performance is to apply a post-processing method. For example, Han et al.~\cite{han2016seq} introduce Seq-NMS to exploit temporal consistency by constructing a temporal graph to link objects in adjacent frames. With a similar idea, Kang et al.~\cite{kang2016object} generate tubelets by combining single image detections through the video and use a tracker to re-score the tubelets during post-processing to improve temporal consistency. 

\subsection{Human Pose Estimation in Videos}\label{sec:related_pose_est}

Most previous work on human pose estimation in videos operates in a single-frame manner~\cite{martinez2017simple}. However, 
utilizing temporal information can be valuable for human pose estimation in videos as it can improve the model robustness and/or efficiency. Using recurrent models~\cite{gkioxari2016chained,lin2017recurrent,artacho2020unipose}, optical flow~\cite{pfister2015flowing,xiao2018simple} and/or temporal convolutions~\cite{pavllo20193d,duan2022revisiting} are some of the common efforts. 

Gkioxari et al.~\cite{gkioxari2016chained} chain predictions in time by utilizing a recurrent network for improved performance. Lin et al.~\cite{lin2017recurrent} accumulate pose information temporally with an LSTM block by feeding the concatenation of the previous pose predictions to regress 3D body joints on the current frame. Artacho et al.~\cite{artacho2020unipose} incorporate an LSTM module into their model and feed it with the heatmaps from the previous video frames along with the current frame's heatmap and generate the heatmap for the final prediction.

Pfister et al.~\cite{pfister2015flowing} generate heatmaps for all frames individually and warp them using optical flow to later pool them on the current frame to make its predictions. Xiao et al.~\cite{xiao2018simple} use optical flow and a tracker that performs greedy matching to shift and track the poses in time. 

Pavllo et al.~\cite{pavllo20193d} use dilated temporal convolutions over a sequence of 2D keypoints to capture long term information and generate 3D keypoint detections. Duan et al.~\cite{duan2022revisiting} uses a two-stage pose estimation where the first step is to detect bounding boxes. Using those bounding boxes, the model estimates poses on each frame. Finally, these predictions are stacked to create 3D heatmap volumes which are then processed with temporal convolutions for classification.

\subsection{Saliency Based Techniques}

To reduce computational cost, another viable approach is to identify and process only the most informative regions or salient objects within an image~\cite{mnih2014ram,ba2015dram,cordonnier2021differentiable,liu2022confidence,liu2023tcgnet,liu2024deep}. We discuss these techniques to draw a conceptual parallel to our work: just as modern saliency-based methods, such as deep unsupervised part-whole relational visual saliency~\cite{liu2024deep} or Type-Correlation Guidance (TCGNet)~\cite{liu2023tcgnet}, achieve efficiency by exploiting \textit{spatial} redundancy and focusing computation on key patches, StreamDEQ achieves efficiency by exploiting \textit{temporal} redundancy and recycling features across time. 

For static images, the region selection process can continue until the model becomes confident about its predictions. However, when applied to videos~\cite{bazzani2011learning,denil2012learning,zhu2018towards,rhee2022distortion,habibian2021skip,chai2019patchwork,bejnordi2022salisa}, such subset selection strategies share shortcomings similar to approaches relying on flow-based intra-frame prediction approximations. The inputs change over time, therefore the models have to choose between relying on optical flow to warp the rest of the features or to omit them entirely, which may result in obsolete representations over time~\cite{zhu2018towards}.

Mnih et al.~\cite{mnih2014ram} and Ba et al.~\cite{ba2015dram} model human eye movements by capturing {\em glimpses} from images with a recurrent structure and process those glimpses at each step. Cordonnier et al.~\cite{cordonnier2021differentiable} propose selecting the most important regions to process by first processing a downsampled version of the image. Liu et al.~\cite{liu2022confidence} stops processing for regions with high-confidence predictions at an earlier stage.

Bazzani et al.~\cite{bazzani2011learning} and Denil et al.~\cite{denil2012learning} approach video processing in a human-like manner where the model {\em looks at} a different patch around the objects of interest at each frame and tracks them. Zhu et al.~\cite{zhu2018towards} take a key frame based approach. At each key frame the method processes the entire input, and at intermediate frames, it updates the feature maps partially based on temporal consistency. Rhee et al.~\cite{rhee2022distortion} identify changing regions between frames and re-use the features of the static parts on non-key frames. Habibian et al.~\cite{habibian2021skip} introduce skip-convolutions where the model determines changing locations via frame difference and computes convolutions only on some part of each frame. Patchwork~\cite{chai2019patchwork} uses a Q-learning based policy to select a sub-window in each frame and combines the sub-window features with the rest of the features via an attention mechanism. SALISA~\cite{bejnordi2022salisa} focuses on an intelligent downsampling method that {\em magnifies} important regions in each frame and reduces the frame's resolution.

\subsection{Implicit Layer Models}\label{sec:related_implicit}

Implicit layer models have seen a recent surge of interest and have been outstanding at tackling numerous tasks. DEQ~\cite{bai2019deq} is a new addition to the implicit model family aimed at solving sequence modeling tasks. DEQ reformulates the fixed point solving problem as root-finding and utilizes an iterative root-finding algorithm to find a solution. Multiscale Deep Equilibrium Models (MDEQ)~\cite{bai2020mdeq} are the extension of the base DEQ to image-based models where there are multiple fixed points for different feature scales. 

Since the introduction of DEQ and MDEQ, there have been many efforts to improve and exploit implicit layer models. Huang et al.~\cite{huang2021impsq} propose re-using the fixed point across training iterations, however with the drawback of having to stay in full-batch mode for the training. Bai et al.~\cite{bai2021neural} suggest a new initialization scheme that is realized through a small network. Furthermore, inferring information from the last few iterations reduces the number of solver iterations required for convergence. Pal et al.~\cite{pal2022mixing} propose mixing implicit models with explicit models. Usually, implicit models are initialized from scratch, i.e. from a random point or zero, to iteratively reach the equilibrium point. Instead, this method first utilizes an explicit model to calculate a ``better'' starting point for the implicit model, resulting in better performance and faster inference. This principle of leveraging a strong initial guess is central to our work. Our work on StreamDEQ adapts this insight to the video domain, showing that the solution from a previous frame serves as an excellent starting point for the next.

Several works have shown the versatility of implicit layer models as a drop-in replacement of their explicit counterparts. Examples of previously explored applications include optical flow estimation~\cite{bai2022deep}, normalizing flows~\cite{lu2021implicit}, graph neural networks~\cite{gu2020implicit}, feature refinement in detection and segmentation pipelines~\cite{ma2021implicit}, multimodal fusion~\cite{ni2023deq_fusion}, Feature Pyramid Networks~\cite{wang2020implicit}, snapshot compressive imaging~\cite{zhao2023deep}, diffusion models~\cite{pokle2022deep,geng2023one}, continuous latent variable estimation~\cite{tsuchida2023deep}, point cloud classification/completion~\cite{geuter2025ddeqs}, and query-based object detection and instance segmentation~\cite{wang2025deqdet}.

Despite their conceptual appeal, DEQs exhibit fundamental limitations that hinder practical deployment. They are brittle meaning minor architectural changes can impair convergence, and training often becomes unstable as it progresses~\cite{bai2021stabilizing}. Moreover, both forward and backward passes require solving fixed point equations with the backward pass being particularly expensive due to the inverse Jacobian computation (Eq.~\eqref{eq:deq_backward_pass})~\cite{geng2021training,fung2022jfb}.

Several works attempt to mitigate these issues. Bai et al.~\cite{bai2021stabilizing} introduce Jacobian regularization to improve stability, while Geng et al.~\cite{geng2021training} and Fung et al.~\cite{fung2022jfb} reduce backward pass cost via truncated Neumann series expansions, showing that low-order inexact gradients suffice in practice.

Another line of work enforces provable convergence guarantees through architectural constraints. Winston and Kolter~\cite{winston2020monotone} and Chen et al.~\cite{chen2021semialgebraic} impose monotonicity to guarantee fixed point existence and uniqueness, while related approaches explicitly bound the Lipschitz constant~\cite{chen2021semialgebraic,el2021implicit,revay2020lipschitz}, often via spectral normalization. Although these methods guarantee convergence, they can restrict expressive power.

\section{Background}~\label{sec:background}

Weight-tied networks are models where some or all layers share the same weights~\cite{bai2018trellis,dehghani2018universal}. A DEQ is essentially a weight-tied network with only one shallow block. DEQ leverages the fact that continuously applying the same layer to its output tends to guide the output to an equilibrium point, i.e. a fixed point. Let $\mathbf{x}$ represent the model's input, $\mathbf{z}^{*}$ the equilibrium point, $f_\theta$ the applied shallow block, and the superscript $[i]$ each iteration, then an explicit weight-tied network can be described as
\begin{align}
    \lim_{i\to \infty}\mathbf{z}^{[i+1]} = \lim_{i\to \infty} f_\theta (\mathbf{z}^{[i]} ; \mathbf{x}) \equiv f_\theta (\mathbf{z}^{*} ; \mathbf{x}) = \mathbf{z}^{*}. \label{eq:fixed_point_iteration}
\end{align}
Each iteration in Eq.~\eqref{eq:fixed_point_iteration} is an unrolling of the model, akin to recurrent neural networks (RNNs), and this scheme is called the fixed point iteration. DEQ's fundamental difference from a standard weight-tied model is that the model is represented by an implicit equation, and the fixed point is found by employing root-finding algorithms in both forward and backward passes, by rewriting Eq.~\eqref{eq:fixed_point_iteration} as follows:
\begin{align}
     g_\theta (\mathbf{z} ; \mathbf{x}) = f_\theta (\mathbf{z} ; \mathbf{x}) - \mathbf{z} = 0 \implies \mathbf{z}^{*} = \text{RootFind}(g_\theta; \mathbf{x}). \label{eq:root_find}
\end{align}
Using Eq.~\eqref{eq:root_find}, the gradients for a given loss $\ell$ can be computed directly only by using the final output \cite{bai2019deq}:
\begin{align}
 \frac{\partial \ell}{\partial (\cdot)} =  \frac{\partial \ell}{\partial \mathbf{z}^{*}}(I-J_{f_{\theta}}(\mathbf{z}^{*}))^{-1}\frac{\partial f_{\theta}(\mathbf{z}^{*};\mathbf{x})}{\partial (\cdot)}, \label{eq:deq_backward_pass}
\end{align}
where $J_{f_{\theta}}(\mathbf{z}^{*})$ is the Jacobian matrix at $\mathbf{z}^{*}$. 

DEQ uses Broyden's method~\cite{broyden1965class} to calculate the root of Eq.~\eqref{eq:root_find}. In this setting, the accuracy of the solution depends on the number of Broyden iterations~\cite{bai2021neural,bai2021stabilizing,ma2021implicit}. While more iterations yield better accuracy, they increase computational cost.

DEQs have been successfully adapted to computer vision tasks with the introduction of Multiscale Deep Equilibrium Models (MDEQ)~\cite{bai2020mdeq}. MDEQ is a multiscale model where each scale is driven to equilibrium together with other scales in the same manner as DEQs. Broyden iterations start with $\mathbf{z}^{[0]}=0$ and continue $N$ times to obtain the final solution, $\mathbf{z}^{[N]}$. The aim is the empirically set $N$ such that $\mathbf{z}^{[N]}=\mathbf{z}^{*}$. In MDEQ~\cite{bai2020mdeq}, $N$ is set to $26$ for ImageNet classification and $27$ for Cityscapes semantic segmentation. 

\section{Proposed Method: Streaming DEQ}\label{sec:method}

In this section, we present the StreamDEQ framework. Before delving into the mathematical details, we provide an architectural overview of our complete pipeline in Fig.~\ref{fig:framework_overview}. This figure illustrates how StreamDEQ actively recycles the equilibrium representation from the previous time step to effectively warm-start the solver for the incoming video frame.

We first introduce our notation and describe the intuition underlying StreamDEQ, as well as the conceptual reasoning that motivated its design. Finally, we conclude this section by introducing the different StreamDEQ variants, i.e. Unrolled and Stochastically Unrolled StreamDEQ.

\begin{figure*}[t]
\centering
\resizebox{0.95\textwidth}{!}{%
\begin{tikzpicture}[
    block/.style={draw, rectangle, rounded corners, minimum width=2.5cm, minimum height=1.2cm, align=center, fill=blue!5, thick},
    memory/.style={draw, rectangle, rounded corners, minimum width=2.5cm, minimum height=1cm, align=center, fill=orange!10, thick},
    head/.style={draw, rectangle, minimum width=2.5cm, minimum height=1.2cm, align=center, fill=green!5, thick},
    arrow/.style={-{Stealth[length=2.5mm]}, thick},
    dashedarrow/.style={-{Stealth[length=2.5mm]}, thick, densely dashed, draw=black!90},
]
\node (input) [align=center] {Input Video Frame \\ $\mathbf{x}_t$};
\node (solver) [block, right=1.2cm of input] {Fixed-Point Solver \\ (Broyden / Unrolled) \\ $\mathbf{z}_t^{[i+1]} = f_\theta(\mathbf{z}_t^{[i]}; \mathbf{x}_t)$};
\node (memory) [memory, above=1.5cm of solver] {Representation Memory \\ $\mathbf{z}_{t-1}^*$};
\node (task) [head, right=2.4cm of solver] {Task-Specific \\ Prediction Head};
\node (output) [align=center, right=1cm of task] {Final Predictions \\ $\mathbf{y}_t$ \\ \small{(Seg., Det., Pose, etc.)}};
\draw [arrow] (input) -- (solver);
\draw [arrow] (memory) -- (solver) node[midway, right, align=left] {Warm-start \\ $\mathbf{z}_t^{[0]} = \mathbf{z}_{t-1}^*$};
\draw [arrow] (solver) -- (task) node[pos=0.5, above=0.05cm] {$\mathbf{z}_t^*$};
\path (solver.east) -- (task.west) coordinate[pos=0.7] (midZ);
\draw [dashedarrow] (midZ) |- (memory.east) node[near start, right=0.15cm, align=left] {Representation \\ Recycling ($t \rightarrow t+1$)};
\draw [arrow] (task) -- (output);
\begin{scope}[on background layer]
    \node (corebox) [draw=black!60, dashdotted, thick, rounded corners=3mm, fill=gray!5, inner sep=0.4cm, fit=(solver) (memory)] {};
    \node [above right, font=\bfseries] at (corebox.north west) {StreamDEQ Framework};
\end{scope}
\end{tikzpicture}%
}
\caption{Overall framework pipeline of StreamDEQ. At each time step $t$, the incoming video frame $\mathbf{x}_t$ is processed by the fixed-point solver. Instead of starting from scratch, the solver is initialized (warm-started) using the equilibrium representation $\mathbf{z}_{t-1}^*$ from the previous frame, retrieved from memory. The resulting representation $\mathbf{z}_t^*$ is then passed to the task-specific prediction head and simultaneously recycled into memory for the next frame.}
\label{fig:framework_overview}
\end{figure*}

\subsection{Notation}\label{sec:notation}

Let $\mathbf{X}$ be an ${H \times W \times 3 \times T}$ dimensional tensor representing a video where $T$ is the temporal dimension. We represent the frame at time $t$ with $\mathbf{x}_t$ which is a $H \times W \times 3$ tensor. Similar to the notation of DEQs, we denote the fixed point of the frame $\mathbf{x}_t$ with $\mathbf{z}^{*}_{t}$, the reference representation of that frame.

\subsection{Building the StreamDEQ Framework}\label{sec:building_deq}

Here, we present our intuition behind designing StreamDEQ and how the StreamDEQ framework evolved. We refer to this as a framework, since it is independent of the solution procedure or the model architecture employed by the DEQs. To avoid any confusion and for conciseness' sake, throughout Section~\ref{sec:building_deq}, we refer to the implicit version of StreamDEQ (IL-StreamDEQ). We describe the unrolled variants of StreamDEQ in Section~\ref{sec:unrolled_deq}.

To process a video with DEQs, the model can be applied to each video frame $\mathbf{x}_t$ individually to obtain $\mathbf{z}_t^{*}$. This amounts to running the solver, e.g. Broyden's solver, starting from $\mathbf{z}_t^{[0]}=\mathbf{0}$ on each frame for $N$ iterations, i.e. until convergence.

\begin{figure}[t]
\centering
\includegraphics[width=\figmulti\linewidth]{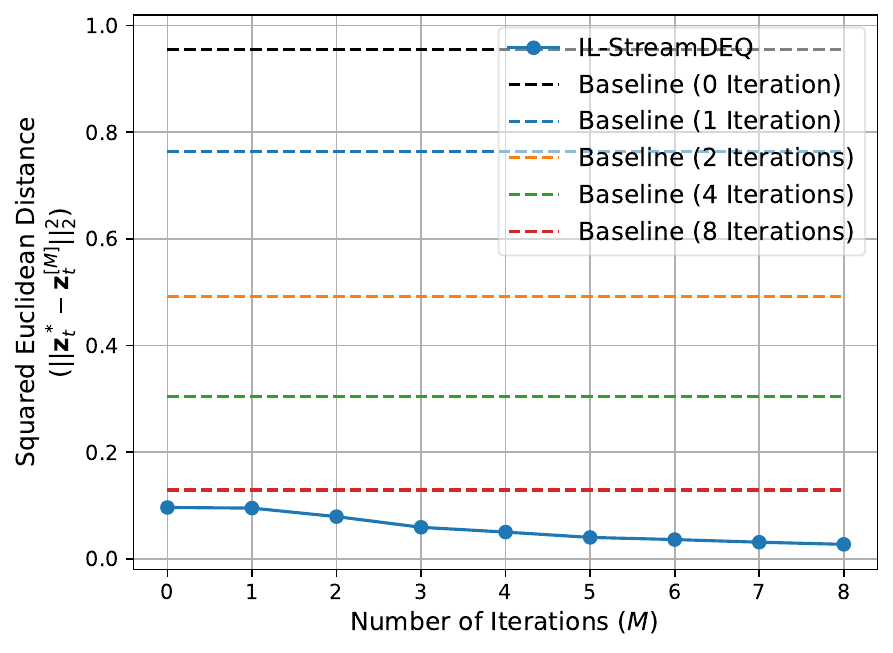}
\caption[Approximation error when the solver is initialized with the reference representation of the preceding frame.]{Squared Euclidean approximation error as a function of inference steps, when the solver is initialized with the reference representation of the preceding frame.}\label{fig:l2vsiter}
\end{figure}

However, it is reasonable to assume a priori that transitions between subsequent video frames are typically smooth, i.e. $\mathbf{x}_{t-1} {\sim} \mathbf{x}_t$. From this observation, we hypothesize that the corresponding fixed points, i.e. representations $\mathbf{z}_{t-1}^{*}$ and $\mathbf{z}_{t}^{*}$, are likely to be similar. Therefore, the representation of the previous frame can be used effectively as a starting point for inferring the representation of the current frame. To validate this hypothesis, we conduct an analysis on the ImageNet-VID~\cite{imagenetviddataset} dataset using the ImageNet pretrained MDEQ model. We assume that at each frame $\mathbf{x}_t$, we have access to the reference representation, $\mathbf{z}_{t-1}^*$, of the previous frame. Reference representations are obtained by running the MDEQ model until convergence, or $N=26$ iterations with the Broyden's solver. At each frame, we use the reference representation of the previous frame as the starting point of the solver, 
\begin{equation}
    \mathbf{z}_t^{[0]} = \mathbf{z}_{t-1}^* , \label{eq:model1}
\end{equation}
\noindent and run the solver for various small numbers of iterations, $M$. To analyze the amount of change in representations over time, we use an ImageNet-pretrained model since ImageNet representations are known to be useful in many transfer learning tasks. In Fig.~\ref{fig:l2vsiter}, we show the squared Euclidean distance between $\mathbf{z}_t^{[M]}$ and $\mathbf{z}_t^*$ for various $M$ values when we initialize the solver as in Eq.~\eqref{eq:model1}. Dashed lines correspond to the squared Euclidean distance between MDEQ's reference and $M$-iteration based representations. Note that the dashed black line indicating the $0$ iterations case shows the distance of the reference representation to the original starting point.

From the results presented in Fig.~\ref{fig:l2vsiter}, we observe that after starting from the reference representation of the previous frame and not performing any iterations, the approximate representation is already more similar to the reference representation than starting from scratch and performing $8$ iterations (the solid blue line starts below the red dashed line even at $0$ iterations). We also observe that when we initialize the solver with the preceding frame's fixed point, the inference process converges towards the reference representation in under $8$ iterations. Our findings noticeably agree with those in Pal et al.~\cite{pal2022mixing}, which show that good initializations of the solver generally lead to better convergence for single images, and indicate that the fixed point of the preceding frame is a good starting point for calculating the fixed point of the following frame in videos. 

\begin{figure}[t]
    \centering
    \includegraphics[width=\figmulti\linewidth]{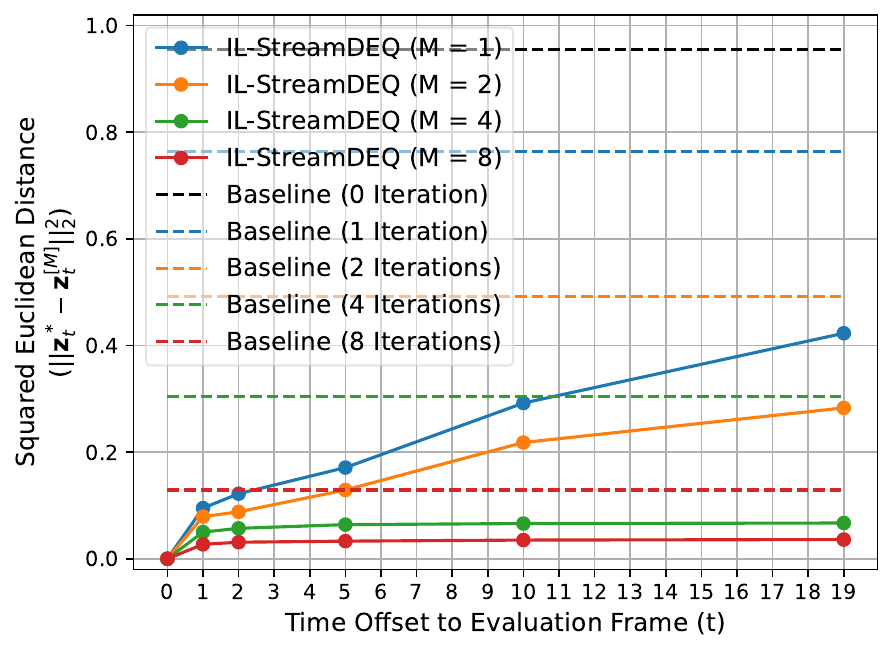}
\caption[Distance between the reference representations and IL-StreamDEQ estimations.]{Distance between the reference representations and IL-StreamDEQ estimations for varying number of iterations, when IL-StreamDEQ is initialized with reference representations on the first frame.}
\label{fig:l2_iter_frame_ref}
\end{figure}

Next, we examine the case where the inference method is given access to the reference representations only at certain frames. To simulate this case, for each video clip, we compute the reference representation at the first frame $\mathbf{x}_0$, i.e. $\mathbf{z}_0^*$. In the subsequent frames, we initialize the solver with the estimated representation of the preceding frame and run the solver for $M$ iterations. That is, 
\begin{equation}
 \mathbf{z}_1^{[0]} = \mathbf{z}_0^* \;\;\mathrm{and}\; \mathbf{z}_t^{[0]} = \mathbf{z}_{t-1}^{[M]} . \label{eq:model2}
\end{equation} 
We present the results of this scheme for $M \in \{1,2,4,8\}$ in Fig.~\ref{fig:l2_iter_frame_ref}. To ensure that the analysis happens on the same fully annotated frame with differing video lengths, we use the following evaluation strategy. Suppose that the test video clip has an annotated frame at time $n$. When we start StreamDEQ's inference at frame $n-t$, run it up to frame $n$ and evaluate it at frame $n$, we record its squared Euclidean distance at the horizontal axis equal to $t$. For example, the point at $t=10$ on the blue curve in Fig.~\ref{fig:l2_iter_frame_ref} corresponds to the squared Euclidean distance obtained when IL-StreamDEQ is started $10$ frames before the evaluation frame. 

We observe that starting with the reference representation on the initial frame is still useful, but for longer clips, its effect diminishes. Still, this scheme helps us maintain a stable performance even after several frames. For example, starting with the reference representation and then applying $M=2$ iterations per frame throughout the following $20$ frames yields a representation closer to the reference representation of the final frame than the one given by baseline DEQ inference with $4$ solver iterations. This result shows that the $M$-step inference scheme is able to keep up with the changes in the scene by starting from a good initial point. 

\begin{figure}[t]
    \centering
    \includegraphics[width=\figmulti\linewidth]{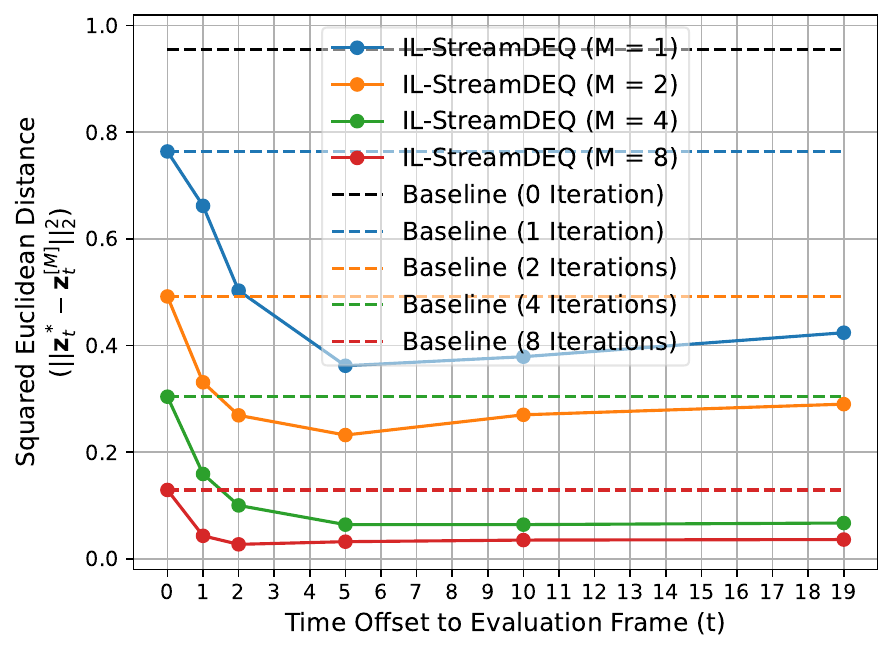} 
\caption[Distance between the reference representations and IL-StreamDEQ estimations, when IL-StreamDEQ is initialized with zeros on the first frame.]{Distance between the reference representations and IL-StreamDEQ estimations, when IL-StreamDEQ is initialized with just zeros on the first frame.}
\label{fig:l2_iter_frame}
\end{figure}

\subsection{StreamDEQ Formulation}\label{sec:streamdeq_formulation}

While the aforementioned scheme can provide efficient inference on novel frames, we would still need the reference representations of the initial frames, or key frame(s), which would share the same problems with key frame based video recognition approaches, e.g.~\cite{zhu2017dff,xu2018dynamic,liu2019looking}. To address this problem, we further develop the idea and hypothesize that we can start from scratch (i.e. all zeros), do a limited number of iterations per frame, and pass the representation to the next frame as the starting point. That is,
\begin{equation}
    \mathbf{z}_0^{[0]} = \mathbf{0} \;\;\mathrm{and}\; \mathbf{z}_t^{[0]} = \mathbf{z}_{t-1}^{[M]} . \label{eq:model3}
\end{equation} 
We present the corresponding representation distance results in Fig.~\ref{fig:l2_iter_frame}. The representation distances to the reference representations stabilize in $20$ frames. The converged distance values (in $20$ frames) are similar to those in the previous scheme (Eq.~\eqref{eq:model2}, Fig.~\ref{fig:l2_iter_frame_ref}). Additionally, the initial representations have relatively large distances, but these differences get smaller as new frames arrive. Hence, we arrive at the StreamDEQ framework. \emph{This scheme avoids heavy processing at any frame and completely avoids the concept of key frames.} The number of iterations can be tuned, which allows controlling the time/accuracy trade-off easily. Therefore, the inference iterations can be run as much as the time budget allows. 

\subsection{Unrolled \& Stochastically Unrolled Streaming DEQ}\label{sec:unrolled_deq}

In addition to using the root-finding formulation, we can also formulate our scheme as the fixed point iteration problem from Eq.~\eqref{eq:fixed_point_iteration}, i.e. the weight-tied setting. This allows us to approach the fixed point solving problem through the more standard CNN-like explicit point of view. This only changes how the fixed point is found while keeping everything else the same, e.g., $f_{\theta}$. We refer to this version as Unrolled StreamDEQ (UR-StreamDEQ). 

The original intuition behind DEQs relied on requiring infinitely many repetitions in the weight-tied setting in Eq.~\eqref{eq:fixed_point_iteration}. With the unrolled version of StreamDEQ, we revisit this assumption by replacing the infinite number of iterations with a finite and small number of iterations. In this setup, we empirically demonstrate that we can match, and even surpass, the downstream performance of infinite depth networks with a very small number of model repetitions, e.g. $20$, instead of relying on black-box root-finding methods such as the Broyden's method (Section~\ref{sec:experiments}). Therefore, in this setting, Eq.~\eqref{eq:fixed_point_iteration} becomes:
\begin{align}
    \mathbf{z}^{[i+1]} = f_\theta (\mathbf{z}^{[i]} ; \mathbf{x}) \equiv f_\theta (\mathbf{z}^{*} ; \mathbf{x}) = \mathbf{z}^{*}, \text{ for } i < 20. \label{eq:unrolled_equation}
\end{align}
This eliminates the need to estimate the high-cost inverse of a large Jacobian matrix~\cite{bai2021stabilizing,geng2021training,fung2022jfb} required during backpropagation (Eq.~\eqref{eq:deq_backward_pass}). The optimizer switches to highly-optimized methods, such as PyTorch AutoGrad~\cite{pytorch}, resulting in reduced training times.

\begin{figure*}
\centering
\includegraphics[width=\figmulti\linewidth]{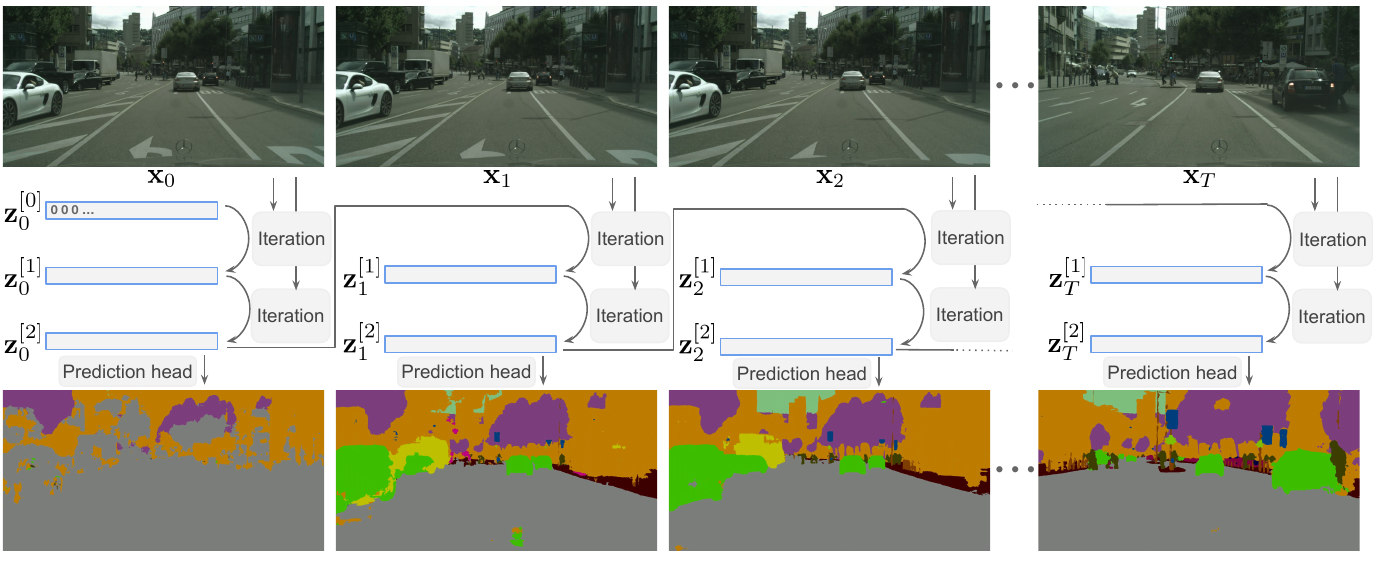}
\caption{StreamDEQ applied to a streaming video, using two iterations per frame. The process is initialized with zeros in the very first frame ($\mathbf{z}_0^{[0]}=\mathbf{0}$) and with the most recent representation ($\mathbf{z}_t^{[0]}=\mathbf{z}_{t-1}^{[2]}$) afterwards. This scheme effectively recycles all recent computations for time-efficient inference and therefore, allows approximating a long inference chain (i.e. a deep network) by a few inference steps (i.e. a few layers) throughout the video stream.}
\label{fig:model_picture}
\end{figure*}

As the only difference between the implicit and explicit versions is the procedure for solving the fixed point equation (Eq.~\eqref{eq:root_find} vs. Eq.~\eqref{eq:unrolled_equation}), our previous arguments regarding the {\em smoothness} of videos and the nature of fixed points of neighboring frames are still valid for UR-StreamDEQ. We do not repeat those experiments here, yet we can still argue that starting with either the reference representations (Eq.~\eqref{eq:model2}) or from scratch (Eq.~\eqref{eq:model3}), UR-StreamDEQ eventually reaches a stable condition where its representations yield comparable performance with that of the reference representation, for the downstream task. We confirm this claim with our experimental findings in Section~\ref{sec:experiments}. While our high-level architecture is outlined in Fig.~\ref{fig:framework_overview}, we present a detailed temporal visualization of the StreamDEQ scheme in action in Fig.~\ref{fig:model_picture}. This illustrates the continuous feature evolution and warm-start initialization across consecutive video frames.

We aim to further improve UR-StreamDEQ by randomizing the number of unrollings during training. Importantly, this modification only affects the training phase: for each batch of training frames, instead of using a fixed number of unrollings (e.g., $20$), we set the number of unrollings uniformly and at random (e.g., between $1$ and $20$), while keeping the test-time inference the same. We refer to this variant as Stochastically Unrolled StreamDEQ (SUR-StreamDEQ).

\begin{figure}
    \centering 
    \includegraphics[width=\figmulti\linewidth]{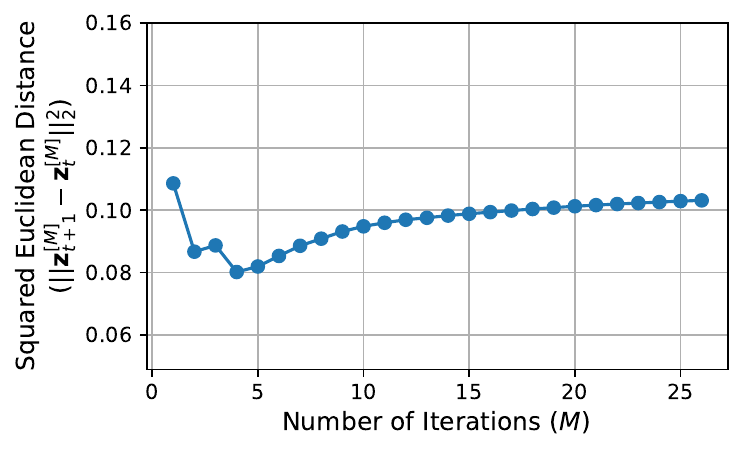}
    \caption{Squared Euclidean distance of two consecutive frames' iteration processes throughout all iterations.}
    \label{fig:l2_consecutive}
\end{figure}

We note that most previous works~\cite{bai2019deq,bai2020mdeq,wang2020implicit,ma2021implicit,bai2022deep} report observations on the advantages of using Broyden iterations over explicit weight-tied versions, typically in terms of achieving quicker convergence. In Section~\ref{sec:experiments}, we empirically demonstrate that this is not always the case and fewer than $20$ unrollings (i.e. fixed point iterations) are sufficient to reach an equilibrium in most scenarios. In most cases, even when the model is unrolled a few times per frame in streaming evaluation, we observe performances on the same level as the variants using black-box root-finding methods, with a noticeably smaller computational burden. We finally note that our video-inference findings are in line with the recent observations on unrolled DEQ inference on images reported in Fung et al.~\cite{fung2022jfb}.

\section{Experimental Results} \label{sec:experiments}

In this section, we first validate our intuition we introduced in Section~\ref{sec:method}. We then show the effectiveness of our method on three streaming video analysis tasks: object detection, semantic segmentation and human pose estimation. For each task, we evaluate all three StreamDEQ variants, denoted with the prefixes {\em IL-}, {\em UR-}, and {\em SUR-}. To remind briefly: {\em IL} uses the Broyden solver; both {\em UR} and {\em SUR} use explicit unrolling (i.e. repeated application of the model on its output), {\em SUR} is trained with a random number of unrollings. In the following, we provide details on the datasets we use, the training and inference setups, and present our results for each task. We conduct all our experiments in PyTorch~\cite{pytorch}. 

\subsection{Confirming Our Intuition}\label{sec:intuition}

To understand the asymptotic behavior of StreamDEQ better, we present three additional empirical results using IL-StreamDEQ. In the first one, we measure the squared Euclidean distance of pairs of consecutive frame representations as found by the Broyden solver for each $M$ value, starting from zero initializations, across the entire dataset.

The results are shown in Fig.~\ref{fig:l2_consecutive}. We first observe that the distance values are unstable for small $M$ values, which is not surprising as we already know that from-scratch DEQ inference requires a {\em sufficiently large} number of iterations. More importantly, for larger $M$ values ($M \geq 15)$, we observe that distance values start to converge towards $0.1$, which is compatible with our observations in Fig.~\ref{fig:l2vsiter}.

\begin{figure}
\centering \includegraphics[width=0.65\linewidth]{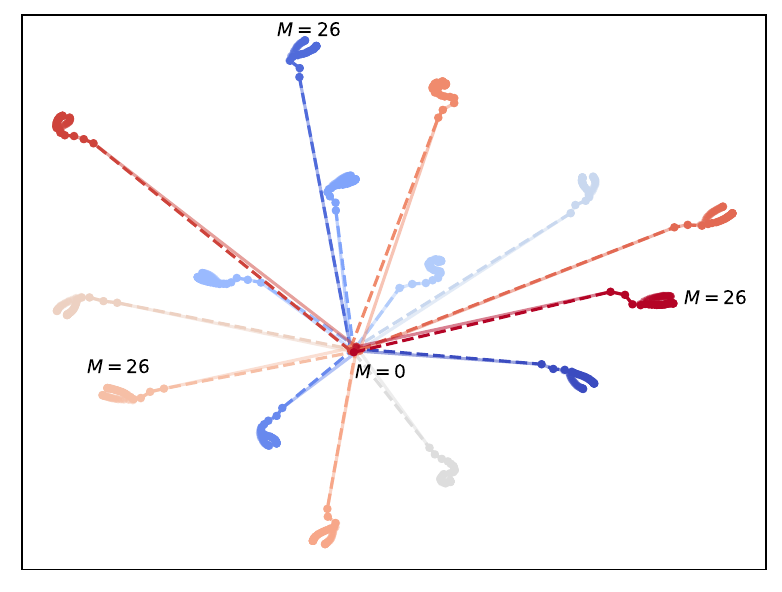}
    \caption{t-SNE visualizations of inference processes. Each color represents a consecutive frame pair. (Best viewed in color)}
    \label{fig:tsne}
\end{figure}

Secondly, we experiment on the similarity of representation inference paths across the consecutive video frames. For this purpose, we investigate whether the consecutive frames converge to fixed points that are in close proximity in the feature space. For this purpose, we visualize the t-SNE~\cite{tsne} embeddings of pairs of consecutive frame representations as found by the Broyden solver for each $M$ value, starting from zero initializations, for $15$ different videos. The results are shown in Fig.~\ref{fig:tsne}. We observe that for $M=0$, all $30$ frames start at a similar location but then, each consecutive frame pair takes a different path. Importantly, we see that each frame in a pair follows a similar trajectory especially during early iterations, e.g., $M \leq 5$ and then due to the differences between consecutive frames, the paths move away from each other while still being widely separated from any other frame pair for $M=26$. This is the idea to which StreamDEQ relies upon indicating that there is strong correlation between consecutive frame pairs' representations during all iterations.

\begin{figure}
\centering
    \includegraphics[width=\figmulti\linewidth]{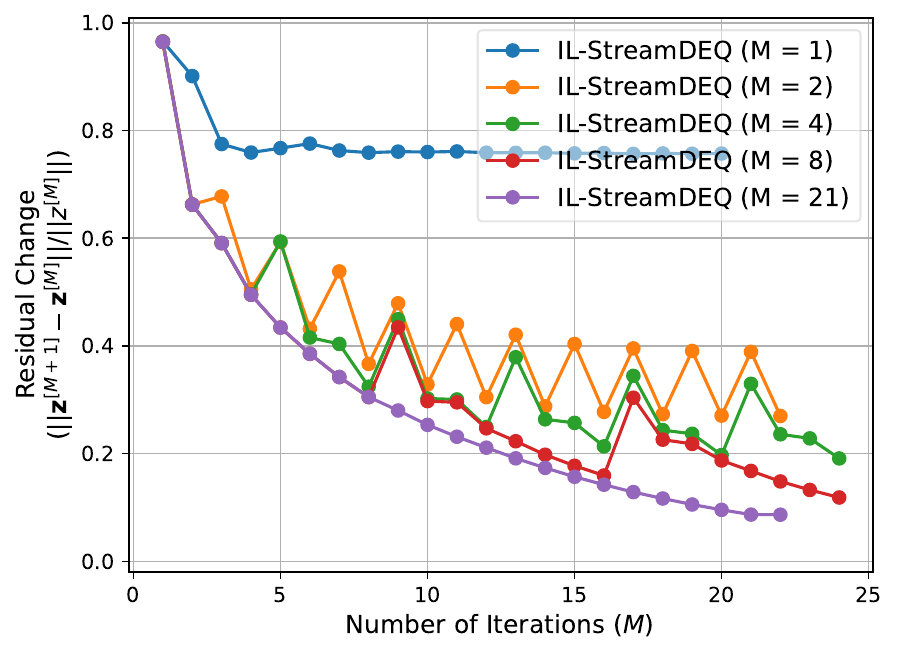}
    \caption{Change of residual per iteration for differing numbers of iterations on streaming videos.}
    \label{fig:rel_residual}
\end{figure}

Lastly, we look into the change of residual ($ \| z^{[M+1]} - z^{[M]}   \| / \| z^{[M]} \| $) for varying $M$ values, inspired from the still image convergence analysis in Bai et al.~\cite{bai2020mdeq}. We perform a similar study on videos using IL-StreamDEQ and present the results for $1, 2, 4, 8,$ and $21$ iterations per frame in Fig.~\ref{fig:rel_residual}. When $M=1$, a frame change occurs at every iteration on the horizontal axis; for $M=2$, a frame change occurs at every second iteration on the horizontal axis and finally, $M=21$ corresponds to the the still image case, as in MDEQ~\cite{bai2020mdeq}. The general trend we see coincides with our intuition given in Fig.~\ref{fig:l2_iter_frame} in the sense that the change of residual drops to lower values as the number of iterations per frame increases, i.e. the number of frames is smaller. For lower numbers of iterations per frame, the trend is still downwards and we observe convergent behavior. Importantly, we clearly see the effect of the frame changes at their respective intervals but even that does not induce a divergent behavior. The points at which a frame change occurs also follow a downward trend, suggesting that the adverse effect of prematurely moving to the next frame is quickly mitigated in the iterations that follow.

\subsection{Video Semantic Segmentation}

\subsubsection{Experimental Setup}

We use the Cityscapes semantic segmentation dataset~\cite{cityscapesdataset}, focusing on its subset of $5$K finely annotated images. The dataset is divided into train, validation, and test sets, each containing $2975$, $500$, and $1525$ images, respectively. They correspond to frames extracted from video clips where each annotated image is the $20^\mathrm{th}$ frame of its respective clip. Because only the $20^\mathrm{th}$ frame of these clips contains the fine pixel-level annotations required for evaluation, our continuous temporal analysis is strictly bounded to this 20-frame window. We emphasize that this is an artifact of the dataset's annotation structure, rather than a point of representation drift or model collapse.

For IL-StreamDEQ, we use the pretrained MDEQ segmentation model from the MDEQ paper~\cite{bai2020mdeq} and do not perform any training. However, for (S)UR-StreamDEQ, an off-the-shelf, pretrained model does not exist. Therefore, we train the models with their respective unrolling settings for 480 epochs with $0.01$ initial learning rate with cosine scheduling following the training setup used in the MDEQ paper~\cite{bai2020mdeq}, initially starting from the ImageNet-pretrained model.  Note that we train the model with still images and do not incorporate temporal information during the training phase.

We follow the evaluation setup and hyperparameters used in MDEQ~\cite{bai2020mdeq}, perform the evaluation on Cityscapes \texttt{val} and report {\em mean intersection over union} (mIoU) results. The architecture of the MDEQ consists of $4$ residual blocks of different scales where each block contains convolution layers, group normalization, and ReLU activation. For further details, we refer the reader to MDEQ~\cite{bai2020mdeq}.

\subsubsection{Results}

\begin{figure}
    \centering
    \includegraphics[width=\figmulti\linewidth]{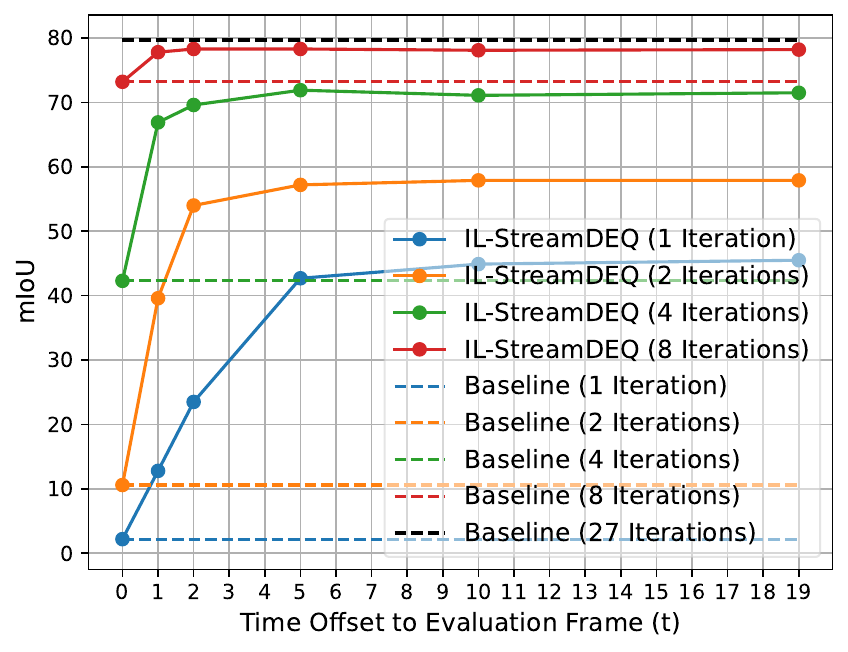} 
\caption[IL-StreamDEQ results on the Cityscapes dataset when the first frame representation is initialized with zeros.]{IL-StreamDEQ semantic segmentation results (in mIoU) on the Cityscapes dataset as a function of solver iterations when the first frame representation is initialized with zeros.}
\label{fig:seg_f_iou_iter}
\end{figure}

We evaluate IL-StreamDEQ using our final StreamDEQ formulation (i.e. Eq.~\eqref{eq:model3}), where we initialize the solver from scratch, i.e. with all zeros, and apply IL-StreamDEQ. The results are shown in Fig.~\ref{fig:seg_f_iou_iter}. As the videos progress, one might expect that the Broyden solver cannot keep up with the changing scenes. However, we observe that even after $20$ frames, the accuracy does not drop. Additionally, the impact of this method is more evident for the lower numbers of iterations. For example, performing $1$ iteration on every frame without our method would only yield an mIoU score of $2.2$. However, IL-StreamDEQ obtains an mIoU score of $44.9$ in $10$ frames, providing an improvement of over $20\times$. For $8$ iterations, IL-StreamDEQ is able to obtain $78.1$ mIoU in $10$ frames, whereas the non-streaming baseline achieves only $73.2$ mIoU. 
Moreover, we also investigate the case where we use the reference representations of the first frame to initialize the solver and apply IL-StreamDEQ then on (Eq.~\eqref{eq:model2}). We observe that the converged mIoU values (at larger frame offsets) are similar in this case compared to Fig.~\ref{fig:seg_f_iou_iter}. This is similar to our observation comparing Fig.~\ref{fig:l2_iter_frame_ref} and Fig.~\ref{fig:l2_iter_frame}. Further details regarding this experiment can be found in the supplementary material. 
Therefore, we conclude that the initial point where we start the solver becomes less crucial as the video progresses and the performance stabilizes at some value higher than that of the non-streaming case.

\begin{figure*}
\centering
\includegraphics[width=\figmulti\linewidth]{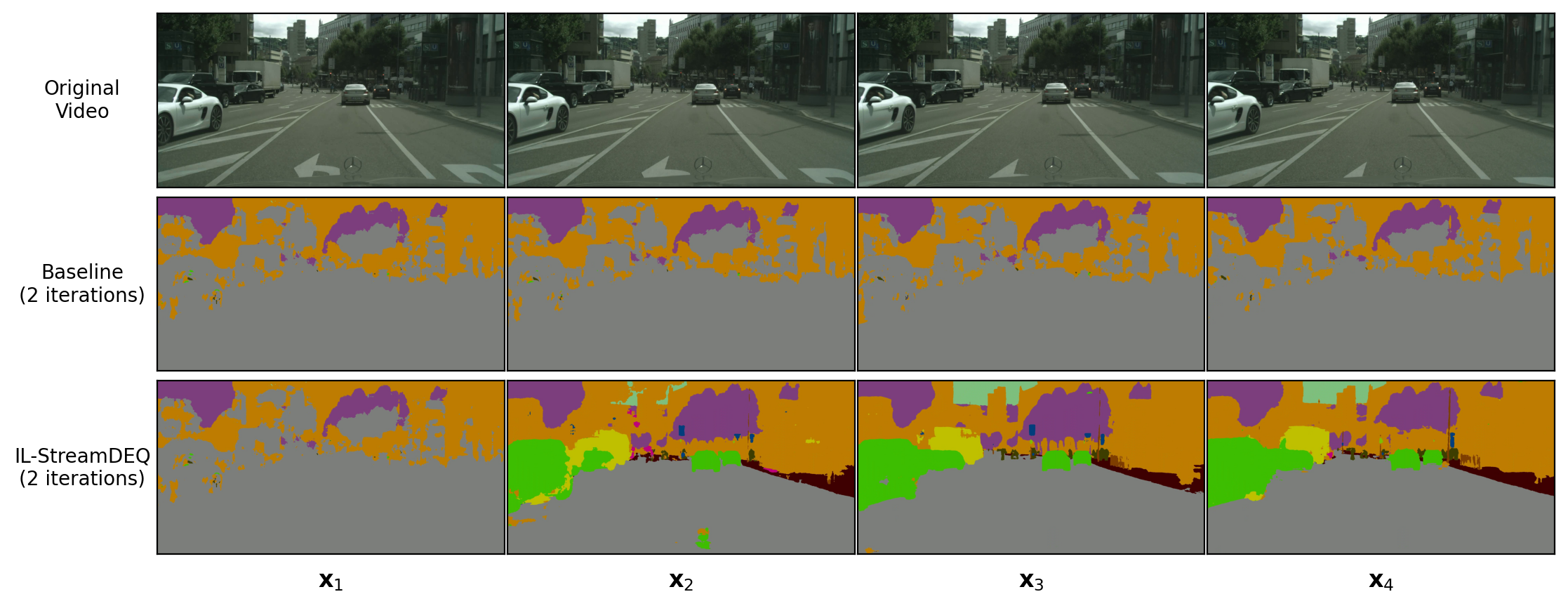}
\caption[Qualitative analysis of StreamDEQ on the Cityscapes dataset.]{Qualitative comparison of the baseline with StreamDEQ with $2$ iterations on the Cityscapes dataset. (Please refer to the supplementary material for a more detailed version of this analysis.)}
\label{fig:seg_qual}
\end{figure*}

We illustrate the results of IL-StreamDEQ qualitatively in Fig.~\ref{fig:seg_qual}. 
With $2$ iterations, while the DEQ baseline produces poor results, IL-StreamDEQ starts to yield accurate predictions as early as the $2^\mathrm{nd}$ frame. Afterwards, IL-StreamDEQ improves its predictions with every new frame producing a clear picture while the DEQ baseline struggles since it has to compute everything from scratch with each frame. We share a more detailed qualitative analysis in the supplementary material (Section~\ref{sec:app_qualitative}).

\begin{figure}
    \centering
    \includegraphics[width=\figmulti\linewidth]{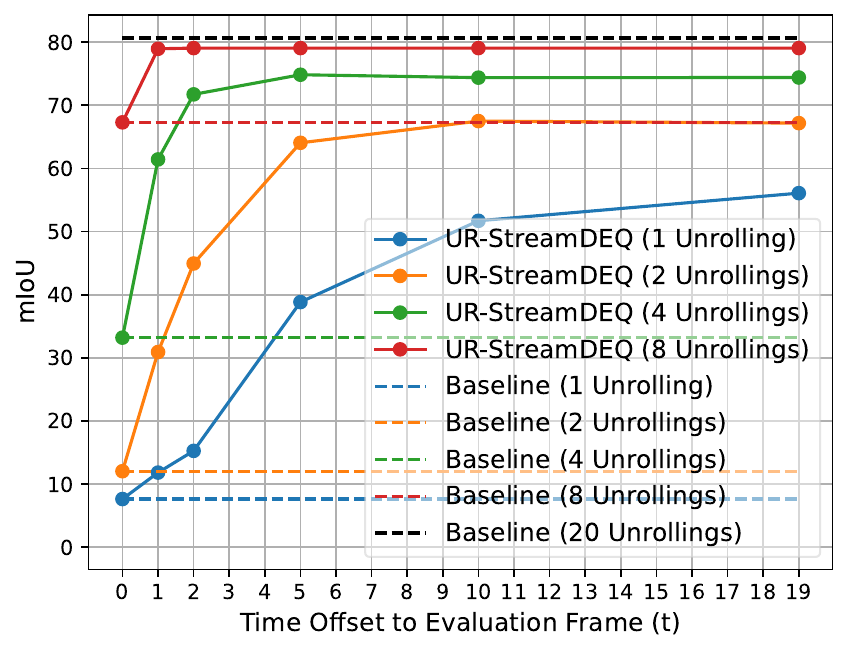} 
\caption[Results of UR-StreamDEQ on the Cityscapes dataset.]{mIoU results of UR-StreamDEQ for various number of iterations after initialization with zeros from the beginning of a clip on the Cityscapes dataset.}
\label{fig:seg_unroll}
\end{figure}

\begin{figure}
    \centering
    \includegraphics[width=\figmulti\linewidth]{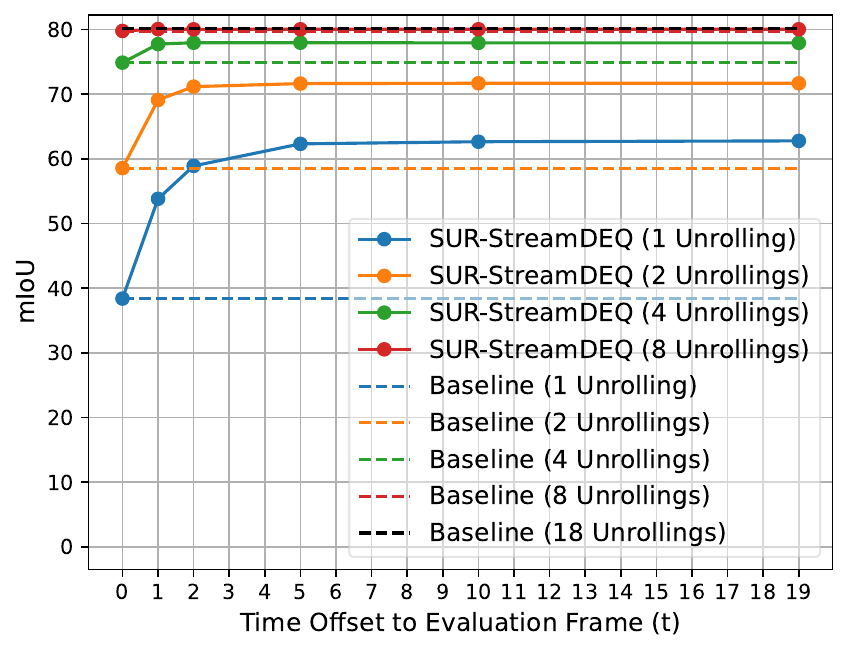} 
\caption[Results of SUR-StreamDEQ on the Cityscapes dataset.]{mIoU results of SUR-StreamDEQ for various number of iterations after initialization with zeros from the beginning of a clip on the Cityscapes dataset.}
\label{fig:seg_unroll_stoch}
\end{figure}

We similarly evaluate the streaming recognition performance of UR-StreamDEQ and SUR-StreamDEQ schemes for varying number of iterations per frame, and present the results in Fig.~\ref{fig:seg_unroll} and Fig.~\ref{fig:seg_unroll_stoch}, respectively. Overall, we observe that (S)UR-StreamDEQ performs better than IL-StreamDEQ. For example, after the first $1-2$ frames, especially SUR-StreamDEQ can reach its performance level that it retains until the end of the video, meaning that it is able to adapt to the changing scenes faster while starting out from $38.4$ mIoU in the $1$ iteration case as opposed to IL-StreamDEQ starting out from $2.2$ mIoU after the first frame. One other observation we have is that in terms of the final performances at the end of the videos, SUR-StreamDEQ comes first (e.g., $71.2$ mIoU for $2$ iterations per frame), followed by UR-StreamDEQ (e.g., $67.2$ mIoU for $2$ iterations per frame), and finally comes IL-StreamDEQ (e.g., $57.9$ mIoU for $2$ iterations per frame). One interesting matter is the performance comparison of UR-StreamDEQ and SUR-StreamDEQ, especially for the lower number of unrollings. We see that SUR-StreamDEQ is again attaining its final performance before UR-StreamDEQ for all numbers of iterations in fewer frames, e.g., for $1$ iteration per frame, after the first frame, UR-StreamDEQ achieves $7.8$ mIoU while SUR-StreamDEQ beats this value by a wide margin achieving $38.8$ mIoU. We attribute this to the model's robustness to the noise introduced by the lower number of iterations seen during training. 

We examine the effects of increasing the number of iterations on inference speed in Table~\ref{tab:inference_speed}. Our method does not introduce any computation overhead other than the time it takes to store the previous frame's fixed point representation. Therefore, we observe a linear increase in compute times as the number of iterations increases. IL-StreamDEQ with $4$ iterations achieves an mIoU score of $71.5$ at $530$ ms per image. MDEQ with $4$ iterations can only achieve $42.3$ mIoU. Even though IL-StreamDEQ performs better than MDEQ, we find that even in the single-frame setting, unrolled variants of StreamDEQ are superior to IL-StreamDEQ both in terms of speed and accuracy as indicated by Table~\ref{tab:inference_speed}. We provide additional inference results with videos on our project page: (\url{https://ufukertenli.github.io/streamdeq/}).

\begin{table*}
\centering
\footnotesize
\begin{threeparttable}
\caption[Inference time and streaming video performance comparisons of StreamDEQ models.]{Performances and inference times of StreamDEQ variants on three streaming video tasks: semantic segmentation, object detection and human pose estimation.}
\label{tab:inference_speed}
\begin{tabular}{cccccccc}
\hline\noalign{\smallskip}
  &  & \multicolumn{2}{c}{Cityscapes} &
      \multicolumn{2}{c}{ImageNet-VID} &
      \multicolumn{2}{c}{MPII} \\
\noalign{\smallskip}
Model & \# iterations$^*$ & mIoU & FPS & mAP@50 & FPS & PCKh & FPS \\
\noalign{\smallskip}
\hline
\noalign{\smallskip}
IL-StreamDEQ & \multirow{3}{*}{1} & 45.5 & 4.3 & 9.1 & 10.3 & 47.9 & 22.3\\\cline{3-8}
UR-StreamDEQ &  & 56.1 & \multirow{ 2}{*}{6.3} & 59.4 & \multirow{ 2}{*}{23.2} & 73.8 & \multirow{ 2}{*}{48.2}\\
SUR-StreamDEQ &  & 62.8 &  & 65.1 &  & 79.6 \\
\hline
IL-StreamDEQ & \multirow{3}{*}{2} & 57.9 & 2.9 & 39.5 & 9.2 & 70.1 & 19.2 \\\cline{3-8}
UR-StreamDEQ &  & 67.2 & \multirow{ 2}{*}{4.9} & 64.1 & \multirow{ 2}{*}{17.4} & 78.6 & \multirow{ 2}{*}{27.4}\\
SUR-StreamDEQ &  & 71.7 &  & 68.8 &  & 86.8 \\
\hline
IL-StreamDEQ & \multirow{3}{*}{4} & 71.5 & 1.9 & 50.4 & 6.2 & 83.6 & 14.4 \\\cline{3-8}
UR-StreamDEQ &  & 74.4 & \multirow{ 2}{*}{3.2} & 65.7 & \multirow{ 2}{*}{11.6} & 86.2 & \multirow{ 2}{*}{22.3}\\
SUR-StreamDEQ &  & 77.9 & & 69.5 & & 89.4 \\
\hline
IL-StreamDEQ & \multirow{3}{*}{8} & 78.2 & 1.1 & 54.8 & 3.5 & 86.2 & 12.2 \\\cline{3-8}
UR-StreamDEQ &  & 79.0 & \multirow{ 2}{*}{1.9} & 67.2 & \multirow{ 2}{*}{6.8} & 84.7 & \multirow{ 2}{*}{19.6}\\
SUR-StreamDEQ &  & 80.0 &  & 70.3 &  & 89.9 \\
\hline
IL-StreamDEQ Baseline$^{**}$  & 27 & 79.7 & 0.3 & 55.0 & 1.2 & 85.6 & 2.7 \\
\hline
UR-StreamDEQ Baseline$^{**}$ &  \multirow{ 2}{*}{20} & 80.6 & \multirow{ 2}{*}{0.9} & 70.8 & \multirow{ 2}{*}{3.1} & 90.4 & \multirow{ 2}{*}{10.6} \\
SUR-StreamDEQ Baseline$^{**}$ &  & 80.2 & & 70.2 & & 90.2 \\
\hline
\end{tabular}
\begin{tablenotes}
\footnotesize
\item[$*$] The number of Broyden iterations for IL-StreamDEQ and number of unrollings for (S)UR-StreamDEQ.
\item[$**$] The marked baselines operate on a single-frame basis.
\end{tablenotes}
\end{threeparttable}
\end{table*}

\subsubsection{Shot Change Experiments}

We also study the effects of {\em shot changes, i.e. camera cuts,} for the video semantic segmentation task using IL-StreamDEQ where we effectively connect two different clips together. Here, we experiment with two different settings. First, we evaluate shot changes within the same context whereby we connect two different videos from the same dataset, i.e. Cityscapes to Cityscapes, as well as an entirely different context where we connect two videos coming from two distinct datasets, i.e. ImageNet-VID to Cityscapes.

\begin{figure}
    \centering
    \includegraphics[width=\figmulti\linewidth]{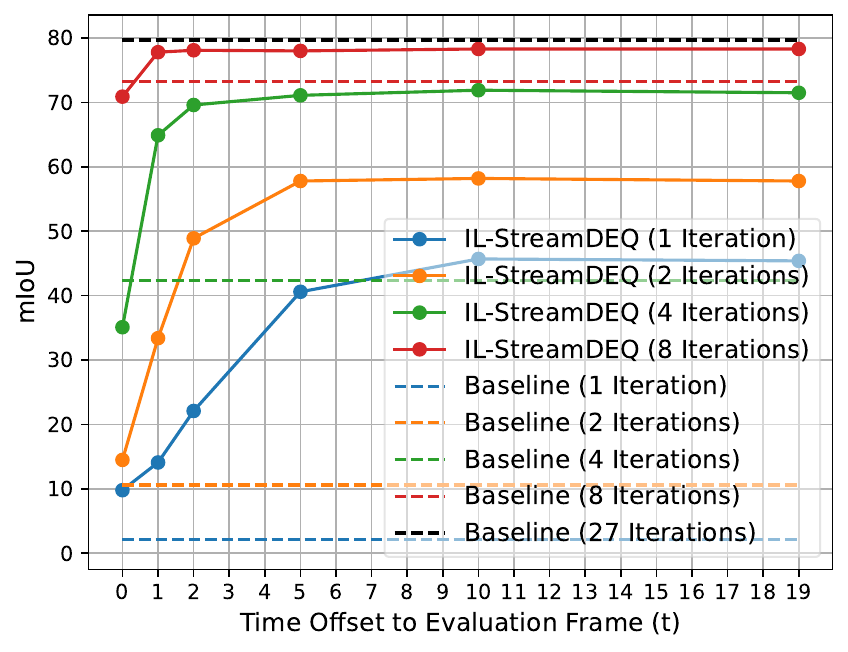}
    \caption[Results of IL-StreamDEQ with shot changes from Cityscapes.]{mIoU results of IL-StreamDEQ with shot changes from the Cityscapes dataset.}
    \label{fig:seg_shot_change_cityscapes}
\end{figure}

\begin{figure}
    \centering
    \includegraphics[width=\figmulti\linewidth]{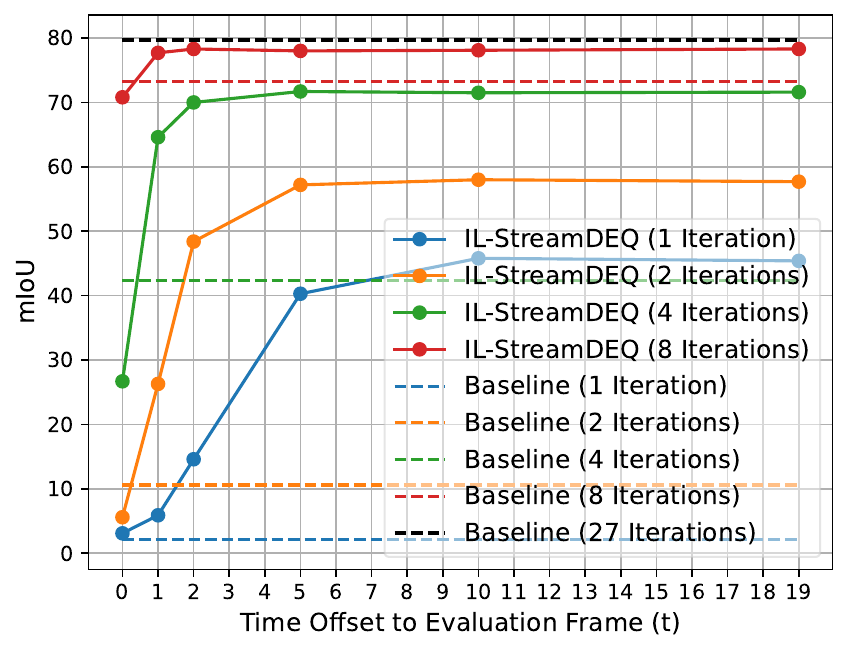}
    \caption[Results of IL-StreamDEQ with shot changes from ImageNet-VID.]{mIoU results of IL-StreamDEQ with shot changes from the ImageNet-VID dataset.}
    \label{fig:seg_shot_change}
\end{figure}

To simulate this behavior, we initialize the solver with the reference representations from a random frame from either the Cityscapes dataset or the ImageNet-VID dataset and run IL-StreamDEQ starting from that representation. We present the results of Cityscapes to Cityscapes shot change experiments in Fig.~\ref{fig:seg_shot_change_cityscapes} and ImageNet-VID to Cityscapes shot change experiments in Fig.~\ref{fig:seg_shot_change}.

The former experiment is simpler, as representations across different videos within the same dataset are likely to be more similar. We notice that, for shot changes in similar contexts, i.e. Cityscapes to Cityscapes, the mIoU scores on initial frames are higher than our previous experiments presented in the main paper in Fig.~\ref{fig:seg_f_iou_iter} and also higher than the ImageNet-VID to Cityscapes shot change scenario in Fig.~\ref{fig:seg_shot_change}. However, after the first few frames are processed, following a similar trajectory to our previous experiments, mIoU scores stabilize at a value close to our original experiment. We conclude that, even with occasional shot changes, our method is able to adapt to the new scene in a few frames.

\subsection{Video Object Detection}

\subsubsection{Experimental Setup}

For the video object detection task, we evaluate our method on the ImageNet-VID dataset~\cite{imagenetviddataset}, a challenging video dataset with fast-moving objects, camera movement, and motion blur. We utilize the MMDetection~\cite{mmdetection} and MMTracking~\cite{mmtracking} frameworks for the implementation. 

The ImageNet-VID dataset consists of $3862$ training and $555$ validation videos from $30$ classes that are a subset of the $200$ classes of the ImageNet-DET dataset. The frames and annotations for each video are available at a rate of $25$-$30$ FPS per video. Note that the ImageNet-DET dataset consists only of still images rather than videos. We follow the widely used protocol~\cite{zhu2017fgfa,wang2018manet,deng2019relation,wu2019selsa,chen2020mega} and train our model on the combination of ImageNet-VID and ImageNet-DET datasets using the $30$ overlapping classes. We use a mini-batch size of $4$, distributed across $4$ NVIDIA A100 GPUs. We resize each image to have a shorter side of $600$ pixels and train the model for a total of $7$ epochs in $3$ stages. We initialize the learning rate to $0.01$ and divide it by $10$ after epochs $2$ and $5$. We test the model on ImageNet-VID \texttt{val} and report mAP@50 scores following the common practice.

We adopt Faster R-CNN~\cite{ren2015faster} by replacing its ResNet backbone with the MDEQ model. To incorporate multi-level representations, we also use a Feature Pyramid Network (FPN)~\cite{lin2017feature} module after MDEQ. Without any additional modifications, we directly utilize the model while keeping the model hyperparameters and remaining architectural details the same as other Faster R-CNN models with ResNet backbones. We only modify the number of channels for the FPN module to match that of MDEQ. We start the training with the ImageNet pretrained MDEQ model~\cite{bai2020mdeq}. 

To train IL-StreamDEQ we use $26$ iterations per frame, following the ImageNet classification experiments in MDEQ~\cite{bai2020mdeq}. (S)UR-StreamDEQ uses the same architecture and training details with the only difference being the solver. We keep the number of unrollings fixed at $20$ for UR-StreamDEQ while SUR-StreamDEQ follows the previous principle whereby we choose a random number between $1$ and $20$ as the number of unrollings during training for each mini batch.

Unlike many video object detection models~\cite{zhu2017fgfa,deng2019relation,chen2020mega}, we train our models in the causal single-frame setting, meaning we do not use any temporal information for improved training. 

\subsubsection{Results}

To the best of our knowledge, this is the first time an implicit model has been used for a video object detection task. We achieve $55.0$, $70.8$, and $70.2$ mAP@50 with IL-StreamDEQ, UR-StreamDEQ, and SUR-StreamDEQ, respectively on ImageNet-VID \texttt{val} (see Table~\ref{tab:inference_speed}). We are aware that Faster R-CNN with ResNet-50 backbone yields $70.7$ mAP@50 off-the-shelf; however, Faster R-CNN is highly optimized to perform well with ResNet backbones. Yet, we use the same setting with an MDEQ without any parameter optimization, as our focus is not on constructing an MDEQ-based state-of-the-art video object detector. We believe there is room for improvement in detector design and tuning details.

\begin{figure}
\centering
    \includegraphics[width=\figmulti\linewidth]{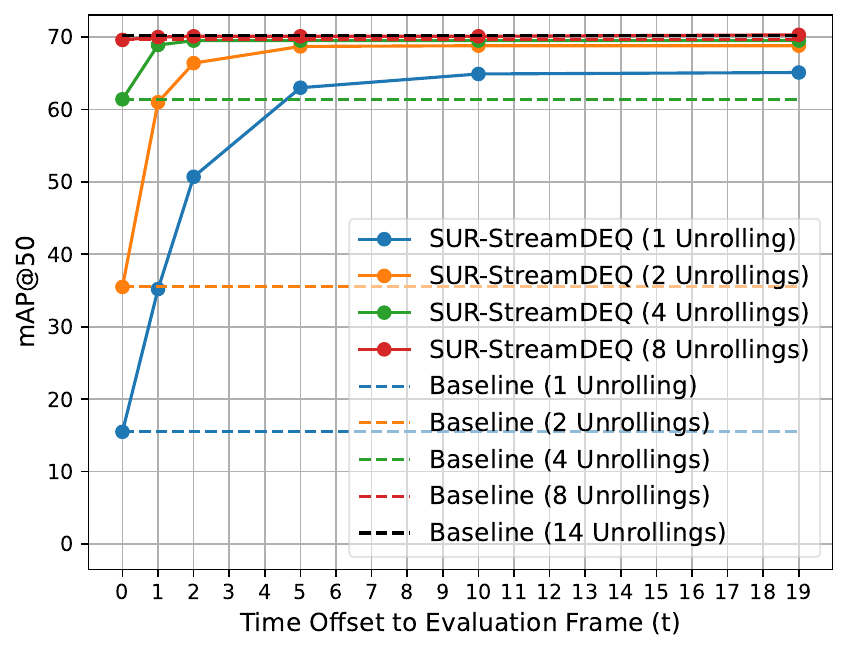}
    \caption[Results of SUR-StreamDEQ on the ImageNet-VID dataset.]{mAP@50 results of SUR-StreamDEQ for various number of iterations after initialization with zeros from the beginning of a clip on the ImageNet-VID dataset.}
    \label{fig:unrollstoch_iter_frame}
\end{figure}

We run IL-StreamDEQ, UR-StreamDEQ, and SUR-StreamDEQ models with different numbers of iterations. We only present the full results obtained with SUR-StreamDEQ and refer readers to Table~\ref{tab:inference_speed} and the supplementary material for the details regarding the other models. We present the results of this experiment in Fig.~\ref{fig:unrollstoch_iter_frame}. 

Our first observation is that the baseline single-frame performances of (S)UR-StreamDEQ are higher than IL-StreamDEQ. Furthermore, even with $1$ unrolling per frame, SUR-StreamDEQ attains about $60$ mAP@50 after $20$ frames. We again discover that the scores stabilize at a value proportional to the number of unrollings. SUR-StreamDEQ starts with a strong $15.5$ mAP@50 even after one frame with $1$ unrolling. We hypothesize that this is due to the robustness to noise we introduce to SUR-StreamDEQ by the randomness in the number of unrollings during training. 

SUR-StreamDEQ is able to sustain its performance with any number of unrollings once the initial frames are processed, e.g. the performance of the model with 2 unrollings per frame reaches to $68.8$ mAP@50 after $10$ frames and does not drop from its maximum value after $20$ frames. Comparing SUR-StreamDEQ with UR-StreamDEQ, we see that the randomness added during training improves the performance which we demonstrate in Table~\ref{tab:inference_speed}. After $20$ frames, SUR-StreamDEQ outperforms UR-StreamDEQ with all numbers of unrollings by $3$-$5$ mAP@50. While UR-StreamDEQ only learns about representations produced with $20$ unrollings per frame, SUR-StreamDEQ has knowledge about a frame's representation after any number of unrollings. Therefore, it is able to adapt to more challenging situations (noisy representations), and its performance improves drastically. Furthermore, SUR-StreamDEQ maintains its performance even after $20$ frames with as low as $2$ unrollings per frame. 

Notice that SUR-StreamDEQ achieves its best single-frame performance using $14$ unrollings. Other StreamDEQ variants use the maximum number of iterations/unrollings that they use during training to make full use of their potential. However, since the number of unrollings per frame is not deterministic with SUR-StreamDEQ, we observe its best single-frame performance using $14$ unrollings, achieving $70.2$ mAP@50. For example, with $20$ unrollings, the score drops to $69.9$ mAP@50. The variations between different numbers of unrollings are minor but still, this shows the effectiveness of the stochasticity by saving $6$ unrollings per frame with a $0.3$ performance improvement even in the single-frame setting.

Additionally, SUR-StreamDEQ catches up to its single-frame performance with both $4$ and $8$ unrollings per frame on streaming videos. Usually, the single-frame performance of any Stream-DEQ model acts as an upper bound to its streaming video performance. This makes sense since the models become accustomed to still images from the training phase, yet, we introduce motion during streaming evaluation. Thus, the performance should degrade slightly due to the injection of moving pictures at each time step. It is interesting to note that SUR-StreamDEQ manages to optimize in such a way that it can tolerate the ``noise'' coming from the shifting objects between successive frames and achieves $70.3$ mAP@50 after $20$ frames. This is $0.1$ more than its single-frame baseline. This demonstrates that even though the performance only marginally improves, with the stochasticity we add to the model during training, the model becomes more effective in the low unrolling scenarios by learning diverse levels of representation complexities.

\subsection{Human Pose Estimation in Videos}\label{sec:pose_est}

\subsubsection{Experimental Setup}

We train and evaluate StreamDEQ on the MPII dataset~\cite{mpiidataset} for human pose estimation in videos. We implement StreamDEQ with the MMPose~\cite{mmpose} framework. 

MPII dataset contains 25k images collected from YouTube videos with over 40k people performing a variety of actions. Even though MPII is an image dataset, it also contains the frames from the YouTube videos, but these frames are unlabeled. We make use of these unlabeled frames when evaluating StreamDEQ on videos, however, only calculate the performance on the annotated frames. While native video pose estimation datasets such as PoseTrack~\cite{andriluka2018posetrack} exist, they introduce complex requirements like multi-person tracking and identity association. We deliberately chose the MPII dataset because it effectively isolates our core evaluation—representation recycling across frames—serving as a bridge between static image evaluation and video sequence simulation without confounding tracking variables.

The architecture consists of the pretrained MDEQ as the backbone and the top down heatmap head from Xiao et al.~\cite{xiao2018simple}. Our overall structure also follows that of Xiao et al.~\cite{xiao2018simple} with MDEQ replacing the backbone. We train the model with the ground-truth bounding boxes, in the single-frame setting without any temporal information for $210$ epochs using a batch size of $256$ with an initial learning rate of $10^{-4}$ which we divide by $10$ after epochs $170$ and $200$. We resize all images to be $256 \times 256$ with our heatmaps being $64 \times 64$. 

All StreamDEQ versions share the same architectural and training details. We train IL-StreamDEQ with $26$ Broyden iterations per frame, UR-StreamDEQ with $20$ unrollings per frame, and finally SUR-StreamDEQ with a random number of unrollings between $1$ and $20$ chosen before processing each mini-batch. We report our results using the Percentage of Correct Keypoints (PCKh@0.5) metric. From here onward, we simply refer to this as PCKh. Furthermore, unlike many common pose detectors~\cite{luvizon20182d,sun2019deep,wang2020deep}, we do not perform flip testing which would improve the test time accuracy but would also increase computational costs.

\subsubsection{Results}

\begin{figure}
\centering
    \includegraphics[width=\figmulti\linewidth]{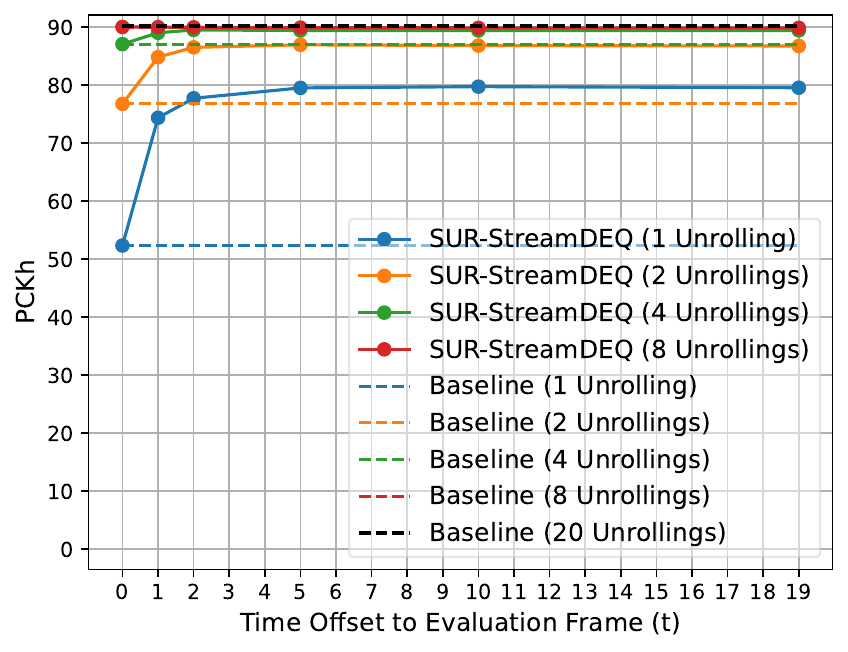}
    \caption[Results of SUR-StreamDEQ on the MPII dataset.]{PCKh results of SUR-StreamDEQ for various number of iterations after initialization with zeros from the beginning of a clip on the MPII dataset.}
    \label{fig:pose_unroll_stoch}
\end{figure}

For this task, we again train the three StreamDEQ variants. We provide an overview of the results in Table~\ref{tab:inference_speed} and the full results for SUR-StreamDEQ in Fig.~\ref{fig:pose_unroll_stoch} and refer readers to the supplementary material for the details regarding IL-StreamDEQ and UR-StreamDEQ. 

The results of this experiment align with our previous experiments, in that, we see the performance of the model improving rapidly after the first few frames. Again due to the stochasticity we introduce to SUR-StreamDEQ, it produces predictions with over $50$ PCKh even after the first frame for all numbers of iterations. Furthermore, performing $4$ iterations per frame versus $8$ iterations per frame only makes a small difference as the performance saturates at about $90$ PCKh for SUR-StreamDEQ. 

Compared to UR-StreamDEQ, SUR-StreamDEQ is able to outperform its deterministic counterpart by $3$-$6$ PCKh during streaming evaluation for any number of iterations (see Table~\ref{tab:inference_speed}) following the pattern in our evaluations with the other downstream tasks.

The model from Xiao et al.~\cite{xiao2018simple} achieves $88.2$ PCKh with a ResNet-50 backbone with flip testing in the single-frame setting~\cite{mmpose}. On the other hand, SUR-StreamDEQ achieves a single-frame performance of $90.2$ PCKh without any test time augmentations. SUR-StreamDEQ outperforms the baseline model in terms of accuracy and speed for both the streaming and single-frame scenarios with streaming performances being superior for more than $16$ PCKh in the $2$ iteration per frame setting while being almost $1.5$ times faster.

\section{Conclusions}\label{sec:conc}

In this paper, we proposed StreamDEQ, an efficient streaming video application of the multiscale implicit deep model, MDEQ. To the best of our knowledge, this is the first large-scale video application of DEQs. Furthermore, we introduced two explicit variants of StreamDEQ. We showed that our models could start from scratch (i.e. all zeros) and efficiently update their representations to reach near-optimal representations as the video streams. We validated this claim on video semantic segmentation, video object detection, and human pose estimation tasks with thorough experiments. 

StreamDEQ models present a viable approach for both real-time video analysis and off-line large-scale methods. StreamDEQ is not specific to segmentation, object detection, or pose estimation and can be used as a drop-in replacement for most other structured prediction problems on streaming videos (action recognition, depth estimation etc.) as long as the prediction task involves an iterative fixed point solution procedure. StreamDEQ works as is in most cases without needing additional tuning demonstrating the versatility of StreamDEQ.

While StreamDEQ offers significant relative speed-ups over single-frame implicit baselines, we acknowledge that its absolute inference speed is currently limited by the relatively unoptimized nature of implicit layer architectures compared to highly tuned explicit backbones (e.g., ResNet). Closing this hardware-level efficiency gap remains an important open problem for the broader implicit modeling research community. However, StreamDEQ is entirely architecture-agnostic, meaning its underlying function ($f_{\theta}$) can be substituted or tuned to fit any criteria. Consequently, the approach proposed in this paper will seamlessly inherit any wall-clock improvements from future developments at the foundational level of implicit model design.

The StreamDEQ framework offers several promising avenues for future research. One key direction is the development of mechanisms to adaptively decide on the number of iterations per frame. Rather than relying on a fixed computational budget, the model could dynamically adjust its effort based on the rate of scene changes, the monitored convergence of features, or via an internal signaling mechanism designed to detect when an equilibrium point has been sufficiently reached. Furthermore, exploring motion-aware StreamDEQ representations could significantly enhance recognition in difficult cases, such as those involving small or fast-moving objects, by more effectively capturing the underlying temporal dynamics. These advancements would further solidify the model's utility in highly dynamic, real-world streaming applications where both efficiency and precision are paramount.

\section*{Acknowledgment}

The numerical calculations were partially performed at TUBITAK ULAKBIM, High Performance and Grid Computing Center (TRUBA) and METU Robotics and AI Technologies Research Center (ROMER) resources. Dr. Cinbis is supported by a Google Faculty Research Award. Dr. Akbas is supported by the BAGEP Award of the Science Academy, Turkey. This work was supported by the Council of Higher Education Research Universities Support program through METU Scientific Research Projects (``New Techniques in Visual Recognition'', Project No. ADEP-312-2024-11485).

\bibliographystyle{IEEEtran}
\bibliography{arxiv}

\clearpage

\appendix

\section{Additional Experimental Results}
\subsection{Overview}

In the supplementary material we present further experimental results on StreamDEQ's convergence and compare IL-StreamDEQ and (S)UR-StreamDEQ on video object detection and human pose estimation in videos. Finally, we provide in-depth qualitative results on video semantic segmentation.

\subsection{Further Empirical Results Confirming Our Intuition on StreamDEQ}

\begin{figure}
    \centering
    \includegraphics[width=\figmulti\linewidth]{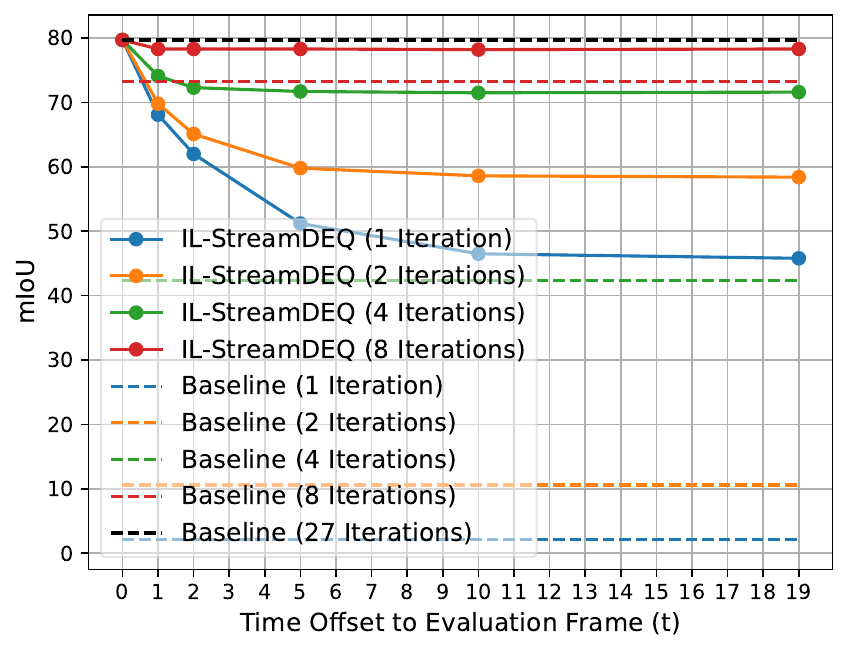}
\caption[IL-StreamDEQ results on the Cityscapes dataset when the first frame representation is initialized with the reference representation.]{IL-StreamDEQ semantic segmentation results (in mIoU) on the Cityscapes dataset as a function of solver iterations when the first frame representation is initialized with the reference representation.}
\label{fig:seg_f_iou_iter_ref}
\end{figure}

Now, we present a comparison of two different initialization schemes for IL-StreamDEQ. We show the first scenario in the main paper that is the original version of IL-StreamDEQ (Eq.~\eqref{eq:model3} from the main paper) where the representations are initially zero and as the videos progress, the representation quality improves. The second scenario corresponds to Eq.~\eqref{eq:model2} from the main paper, where we use the reference representations of the first frame to initialize the solver and apply IL-StreamDEQ then on. We shared the results of these experiments in terms of the squared Euclidean distances in Fig.~\ref{fig:l2_iter_frame} and Fig.~\ref{fig:l2_iter_frame_ref}, respectively. Here, we perform the same experiment on the Cityscapes dataset in terms of the downstream task's evaluation metric which in this case is mIoU. Results of this experiment in Fig.~\ref{fig:seg_f_iou_iter_ref} show that as the offset of the evaluated frame increases, mIoU starts decreasing, which is expected (Fig.~\ref{fig:l2_iter_frame_ref}) because the further we move away from the first frame, the more irrelevant its representation will become. However, mIoU then stabilizes at a value proportional to the number of Broyden iterations (the more iterations, the better the mIoU). This shows that IL-StreamDEQ is able to maintain good quality representations over time even as we move away from the guidance of a reference representation. IL-StreamDEQ's performance with $8$ iterations is still comparable with the baseline (MDEQ) with $27$ iterations. Comparing this experiment to the original IL-StreamDEQ scheme in the main paper, the convergence of mIoU scores towards similar mIoU values (Fig.~\ref{fig:seg_f_iou_iter} from the main paper vs. Fig.~\ref{fig:seg_f_iou_iter_ref}) suggests again that the effect of the first frame's representation diminishes as we move away from the initial point.

\subsection{Comparison Between IL-StreamDEQ and (S)UR-StreamDEQ}

\begin{figure}
    \centering
    \includegraphics[width=\figmulti\linewidth]{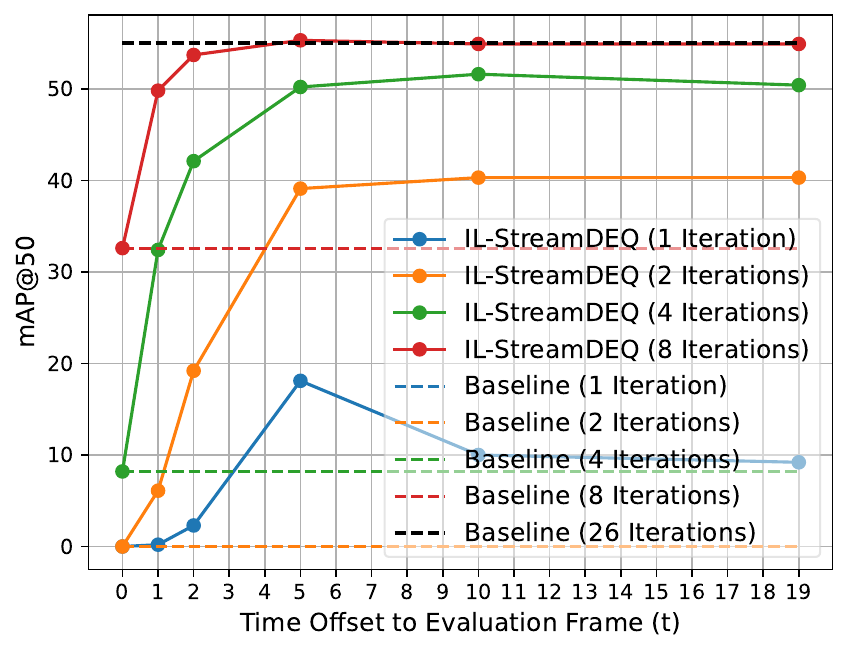}
    \caption[Results of IL-StreamDEQ on the ImageNet-VID dataset.]{mAP@50 results of IL-StreamDEQ for various number of iterations after initialization with zeros from the beginning of a clip on the ImageNet-VID dataset.}
    \label{fig:map_iter_frame_26}
\end{figure}

\begin{figure}
    \centering
    \includegraphics[width=\figmulti\linewidth]{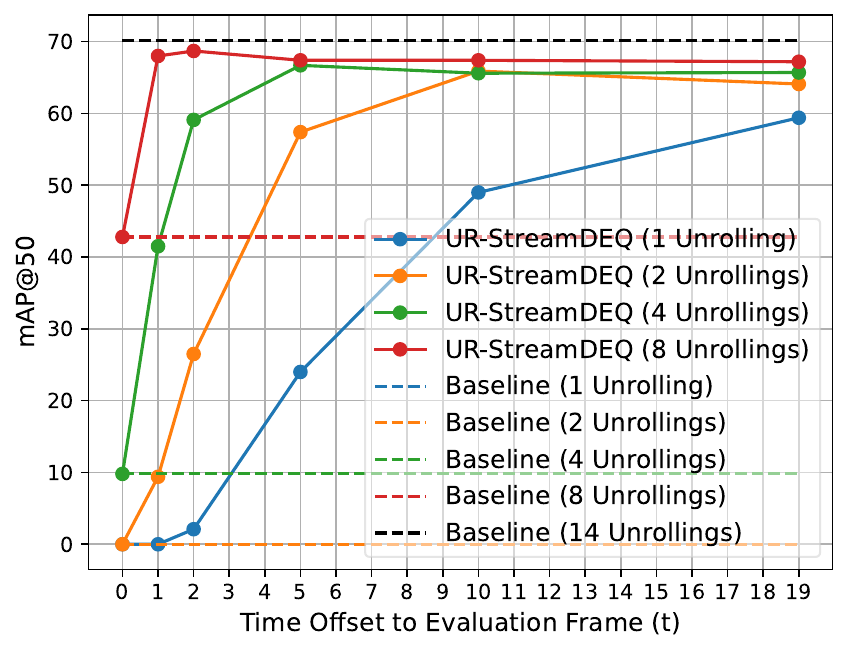}
    \caption[Results of UR-StreamDEQ on the ImageNet-VID dataset.]{mAP@50 results of UR-StreamDEQ for various number of iterations after initialization with zeros from the beginning of a clip on the ImageNet-VID dataset.}
    \label{fig:unroll_iter_frame}
\end{figure}

We present the results of IL-StreamDEQ and UR-StreamDEQ on video object detection in Fig.~\ref{fig:map_iter_frame_26} and Fig.~\ref{fig:unroll_iter_frame}, respectively. We observe the same trends with the segmentation task from the main paper. Over time, the detection performance increases and stabilizes at a value proportional to the number of Broyden iterations or unrollings. 

We note that neither IL-StreamDEQ nor UR-StreamDEQ cannot produce any detection results in the non-streaming mode with $1$ or $2$ iterations. In streaming mode, if we perform $2$ iterations with IL-StreamDEQ, we improve the performance from $0$ to $39.5$ mAP@50 in $20$ frames. 

Again, SUR-StreamDEQ outperforms UR-StreamDEQ for lower number of unrollings. UR-StreamDEQ does not perform well initially for $1$ unrolling per frame and is only able to produce good representations after ${\sim}10$ frames. This is closer to what we see with IL-StreamDEQ in terms of performance. 

\begin{figure}
    \centering
    \includegraphics[width=\figmulti\linewidth]{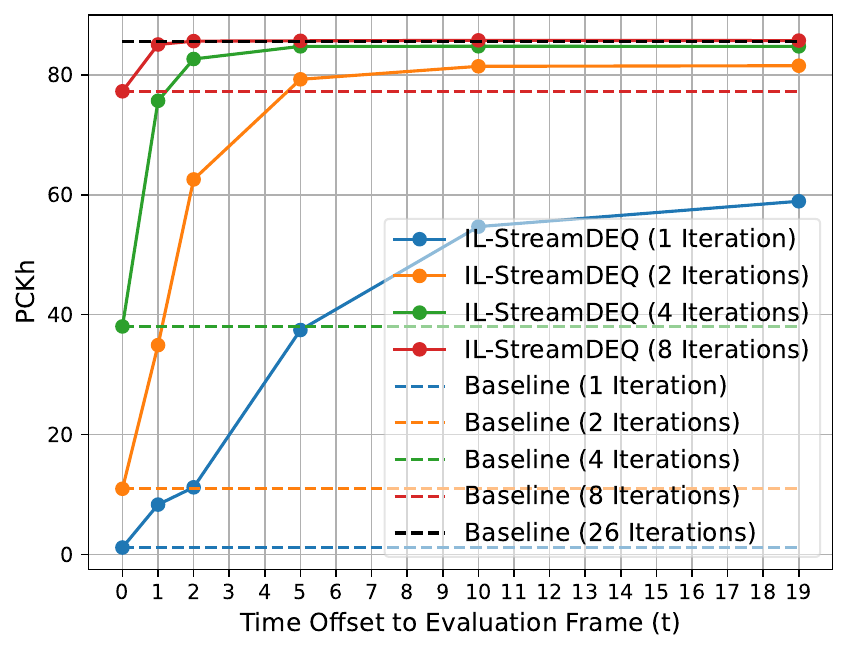}
 \caption[Results of IL-StreamDEQ on the MPII dataset.]{PCKh results of IL-StreamDEQ for various number of iterations after initialization with zeros from the beginning of a clip on the MPII dataset.}
    \label{fig:pose_broyden_26}
\end{figure}

\begin{figure}
    \centering
    \includegraphics[width=\figmulti\linewidth]{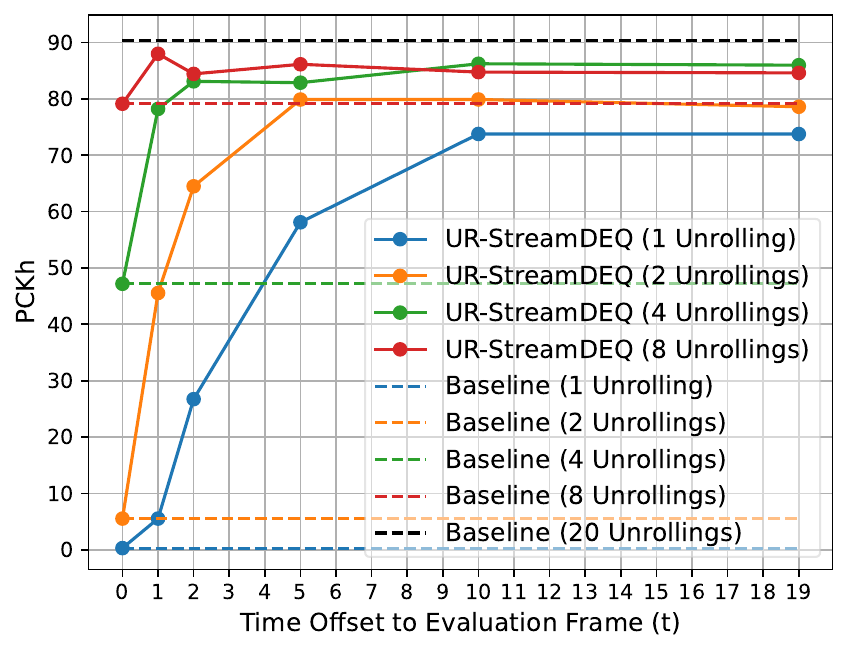}
  \caption[Results of UR-StreamDEQ on the MPII dataset.]{PCKh results of UR-StreamDEQ for various number of iterations after initialization with zeros from the beginning of a clip on the MPII dataset.}
    \label{fig:pose_unroll}
\end{figure}

We also experiment using IL-StreamDEQ and UR-StreamDEQ to estimate human poses in videos whose results we give in Fig.~\ref{fig:pose_broyden_26} and Fig.~\ref{fig:pose_unroll}, respectively.

The results of these experiments again align with our previous experiments, in that, for all $3$ versions of StreamDEQ, we see that the performance of the model improves rapidly after the first few frames. Overall, IL-StreamDEQ performs the worst and SUR-StreamDEQ performs the best. Still, IL-StreamDEQ obtains $84.7$ PCKh and $85.7$ PCKh with $4$ and $8$ iterations per frame, respectively, matching its single-frame performance of $85.7$ PCKh.

UR-StreamDEQ brings an advantage for low numbers of iterations compared to IL-StreamDEQ with its $1$ iteration per frame performance being $73.8$ PCKh. However, we see a slight drop of performance when we perform $8$ iterations per frame after the first couple of frames. We made a similar observation in Fig.~\ref{fig:unroll_iter_frame} and we again suggest that this occurs due to UR-StreamDEQ not having seen representations of different complexities and therefore, is not able to adapt perfectly when it encounters a situation where too many iterations are performed. Yet, its performance drop is not significant where it is able to almost match the performance of the $4$ iterations per frame case ($86.0$ PCKh) at $84.6$ PCKh.

Again due to the stochasticity we introduce to SUR-StreamDEQ, it produces predictions with over $50$ PCKh after the first frame for all numbers of iterations. Furthermore, performing $4$ iterations per frame versus $8$ iterations per frame only makes a small difference as the performance saturates at about $90$ PCKh for SUR-StreamDEQ. Finally, even when we perform $1$ iteration at each frame, SUR-StreamDEQ obtains $79.6$ PCKh which is comparable with the performance of the other StreamDEQ versions when we perform $2$ iterations at each frame. 

\subsection{Qualitative Results}\label{sec:app_qualitative}

\begin{figure*}
    \centering
    \includegraphics[width=\figmulti\linewidth]{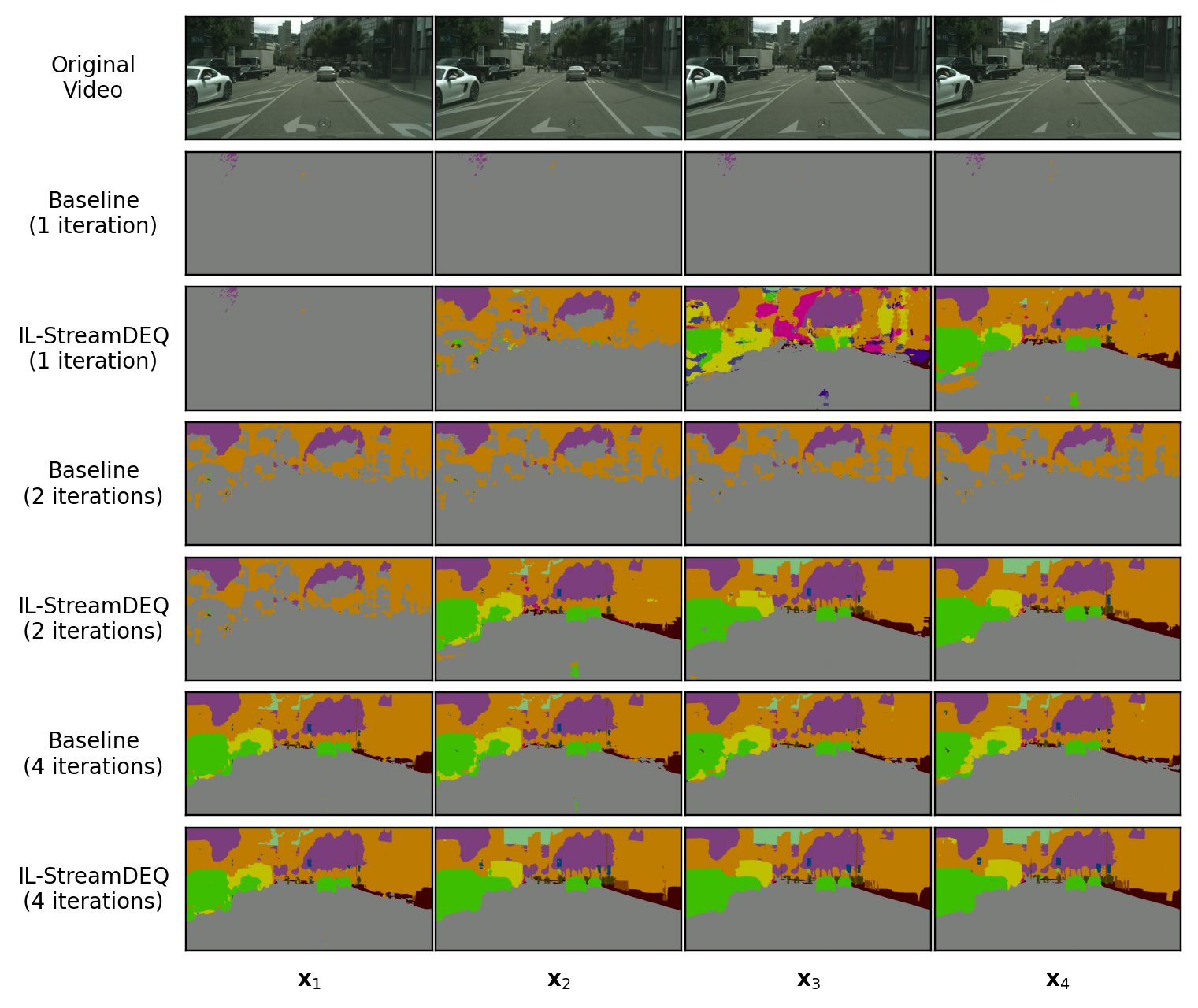}
    \caption[Qualitative analysis of IL-StreamDEQ on the Cityscapes dataset.]{Qualitative comparison of the baseline with IL-StreamDEQ with different numbers of iterations on the Cityscapes dataset.}
    \label{fig:seg_qual_supp}
\end{figure*}

Finally, we illustrate the results of IL-StreamDEQ qualitatively in Fig.~\ref{fig:seg_qual_supp} more in-depth, considering different numbers of iterations, expanding upon Fig.~\ref{fig:seg_qual} from the main paper. For $1$ iteration, while the baseline cannot produce any meaningful segmentations, IL-StreamDEQ starts capturing many segments correctly at the $4^\mathrm{th}$ frame. With $2$ iterations, while the DEQ baseline still produces poor results, IL-StreamDEQ starts to yield accurate predictions in early frames compared to the single iteration case. With $4$ iterations, while both models provide rough but relevant predictions in the first frame, IL-StreamDEQ predictions start to become clearly more accurate in the following frames; for example, tree trunks and the sky becomes visible only with the StreamDEQ scheme.

\end{document}